\documentclass{arxiv_banner}

\paperlogo[11mm]{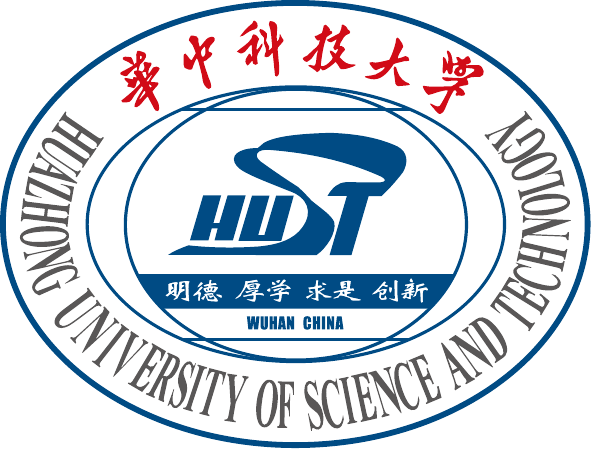}
\paperlogo[11mm][0.05mm]{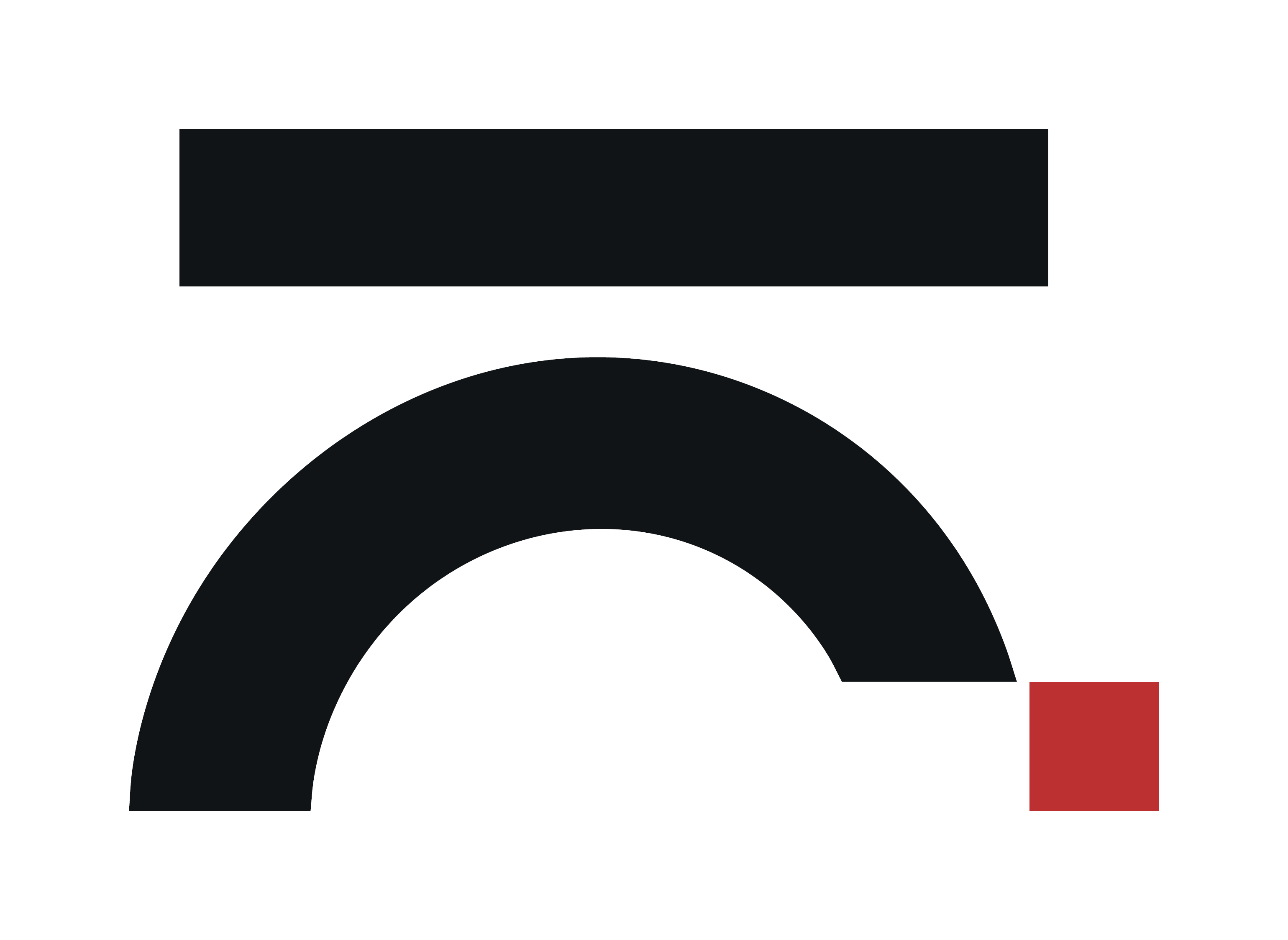}
\paperrightlogo[7.5mm]{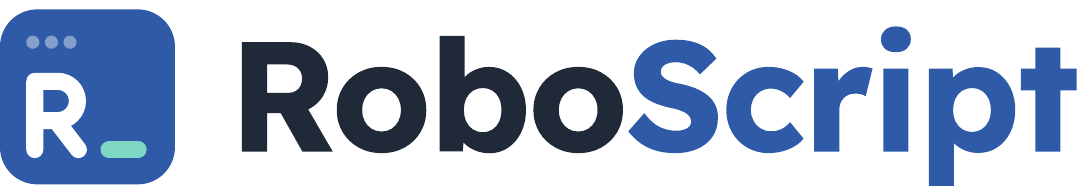}

\title{World In-Context Learning for Agentic \\Robot Manipulation}

\author[1]{Heng Fang}
\author[1]{Zehui Guo}
\author[1]{Yuehan Zhang}
\author[1,2]{Mingyang Du}
\author[1,2]{Dingkang Liang}
\author[1]{Xiang Bai}

\affiliation[1]{Huazhong University of Science and Technology}
\affiliation[2]{ArcheOne AI}
\contact{\{hengfang,dkliang,xbai\}@hust.edu.cn}

\abstract{
Trained action policies write manipulation skill into network weights, tying what is learned to the distribution it is learned from.
In-context learning (ICL) avoids this coupling but has not matched trained policies, because a demonstration recorded in another scene carries no feedback from the world the robot currently faces.
We observe that what an agent learns in context is semantic, recording task steps and object roles rather than visual appearance, so experience gathered in one world transfers to another.
Based on this insight, we propose World In-Context Learning (WICL), a paradigm in which a frozen model starts from a demonstration, rehearses the task in a world such as a simulator, and writes the feedback into its context rather than its weights.
We build \method, a modular infrastructure that implements WICL for agentic manipulation.
From a single demonstration, \method learns each task through rehearsal in simulation, freezes the learned knowledge as a Task Memory, and deploys the frozen memory on new scenes with data isolation that ensures every deployment input is available on a real robot.
On all 50 tasks of RoboTwin~2.0, \method reaches 83.4\% mean success across clean and randomized settings with one demonstration per task and no gradient updates, exceeding the strongest trained action policy by over 20 points.
The method generalizes across visual distributions, improves through reflection, scales with the foundation model, and transfers to a physical dual-arm robot through cross-embodiment memory adaptation.
With a world in the loop, a robot can learn a manipulation task in its context as well as it learns one in its weights.
}
 
\code{https://github.com/ArcheOne-AI/RoboScript}
\project{https://ArcheOne-AI.github.io/RoboScript}

\newcommand{\method}{RoboScript\xspace}

\renewcommand{\figref}[1]{Fig.~\ref{#1}}
\renewcommand{\tabref}[1]{Tab.~\ref{#1}}
\renewcommand{\eqnref}[1]{Eq.~\eqref{#1}}
\makeatletter
\renewcommand{\algref}[1]{\@ifnextchar\bgroup{\ax@algline@ref{#1}}{Alg.~\ref{#1}\xspace}}
\makeatother
\crefname{figure}{Fig.}{Figs.}
\Crefname{figure}{Fig.}{Figs.}
\crefname{table}{Tab.}{Tabs.}
\Crefname{table}{Tab.}{Tabs.}
\crefname{equation}{Eq.}{Eqs.}
\Crefname{equation}{Eq.}{Eqs.}
\crefname{section}{Sec.}{Secs.}
\Crefname{section}{Sec.}{Secs.}
\crefname{subsection}{Sec.}{Secs.}
\Crefname{subsection}{Sec.}{Secs.}
\crefname{subsubsection}{Sec.}{Secs.}
\Crefname{subsubsection}{Sec.}{Secs.}
\crefname{algorithm}{Alg.}{Algs.}
\Crefname{algorithm}{Alg.}{Algs.}

\usepackage{wrapfig}

\begin{document}
\maketitle

\vspace{-1cm}
\section{Introduction}
\begin{figure}[t]
  \centering
  \includegraphics[width=\textwidth]{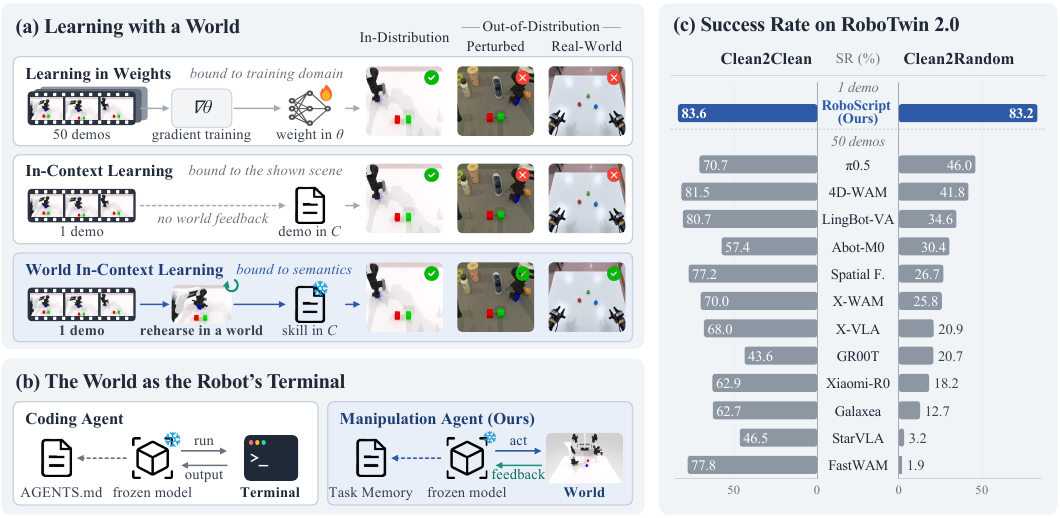}
  \caption{(a) Three paradigms for learning manipulation. Skill in weights is bound to the training domain. ICL avoids retraining but is bound to the shown scene. WICL rehearses in a world and stores semantic knowledge that transfers across settings. (b) \method maps the coding-agent stack to robot manipulation, with the world replacing the terminal. (c) Success rate on all 50 RoboTwin~2.0 tasks with one demonstration and no gradient updates.}
  \label{fig:teaser}
  \vspace{-0.4cm}
\end{figure}
 
In-context learning (ICL) is a defining capability of foundation models.
A model with frozen weights adapts to a new task from what its context shows, without any gradient update \citep{brown2020language,dong2023icl}.
Robot manipulation has long followed a different path.
The dominant approach trains an action policy, either a vision-language-action (VLA) model \citep{black2024pi0,black2025pi05,bjorck2025gr00t,liu2024rdt,kim2024openvla,zitkovich2023rt2} or a world action model (WAM) \citep{ye2026dreamzero,chen2026flowwam}, and writes manipulation skill into network weights.
This approach has made clear progress, but it ties what is learned to the distribution it is learned from.
FastWAM \citep{yuan2026fastwam}, a widely adopted WAM trained on clean demonstrations of RoboTwin~2.0, reaches 77.8\% mean success on clean scenes but drops to 1.9\% under domain randomization \citep{chen2025robotwin}.
Every new task or distribution requires new data and new training.
ICL has recently emerged as a promising alternative that avoids this coupling.
Recent models can perform a new manipulation task from a single demonstration given in their context, without any weight update \citep{generalist2026gen15,skild2026s1,zhou2026zerowam}.
This progress raises the question we study.
Can a robot learn a manipulation task in its context as well as it learns one in its weights?

The in-context route avoids this coupling, but it has not matched trained policies, because its context carries no feedback from the world.
A demonstration is recorded in another scene and cannot tell the robot what will happen on the table in front of it.
A frontier model (e.g.\ GPT-6 Astra) did not improve when given a demonstration from a different layout in its prompt, although some of its traces show it correcting its actions after observing their outcomes \citep{zhang2026unexpected}.
Code as Policies and its successors can control a robot through APIs, but without world feedback their errors accumulate \citep{liang2023codeaspolicies,huang2023voxposer,singh2023progprompt,yu2023languagerewards}.
Coding agents show what is missing.
They learn a new repository by running code in a terminal and reading what comes back \citep{yang2024sweagent,wang2024openhands}.
The feedback loop they rely on has no counterpart in robot manipulation.

Our key observation is that what an agent learns in context is semantic.
It records object roles, task steps, and what went wrong, not the visual appearance of the world.
Experience gathered in one world therefore carries over to another, as long as the two worlds agree on the semantics of the task.
A cheap world that is faithful to the task, such as a simulator, a learned world model, or a latent model of the scene, is enough to learn in.
Rehearsal in such a world is also safe, unlike acting directly on real hardware, where unsafe actions have damaged equipment \citep{zhang2026unexpected}.

We formalize this insight as \textbf{World In-Context Learning} (WICL), a learning paradigm that extends in-context learning with feedback from a world, as illustrated in \figref{fig:teaser}a.
A WICL learner keeps its weights frozen, starts from a demonstration, rehearses the task in a world, and distills the feedback into its context.
The resulting knowledge is semantic and transfers from the rehearsal world to the deployment world without retraining.
After rehearsal, the context is frozen and evaluated on instances that were never rehearsed, so the result measures generalization, not memorization.
The context is readable text, which allows the agent to review its own evaluations and rewrite what it has learned, enabling recursive self-improvement (RSI) without gradient updates.
The world makes every round of improvement safe.
Coding agents have shown that this loop already succeeds in software engineering, where the terminal serves as the world \citep{yang2024sweagent,wang2024openhands}.
Now that foundation models have reached this capability, a simulator can serve the same role for a robot.

\begin{figure}[t]
  \centering
  \includegraphics[width=\textwidth]{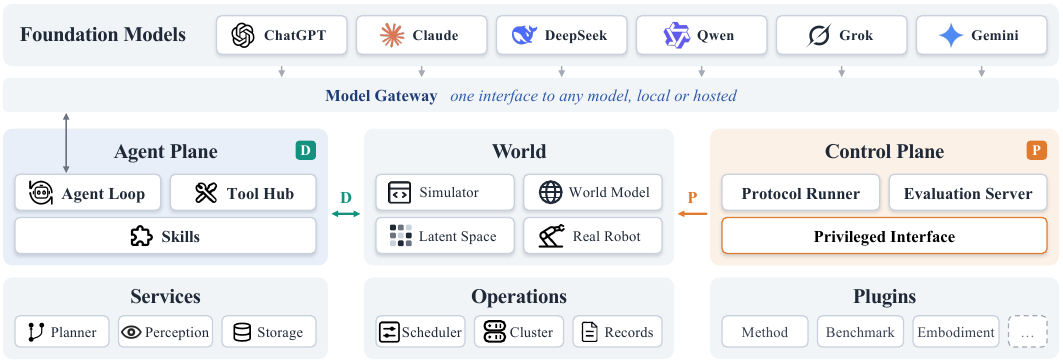}
  \caption{Infrastructure of \method. Every component runs as an independent service, and any method, benchmark, embodiment, or extension plugs in as a replaceable module. All model calls pass through one gateway. The agent plane reaches the world only through the deployable interface (D), and only the control plane accesses privileged data (P).}
  \label{fig:infra}
  \vspace{-0.4cm}
\end{figure}
 We build \textbf{\method}, a modular infrastructure for agent-centric robot manipulation. \figref{fig:teaser}b illustrates the analogy and \figref{fig:teaser}c reports the main results.
In this paper, we implement WICL on \method with a lifecycle that the agent carries out entirely without gradient updates.
The foundation model first builds skills for a new environment on shared entry points.
It then learns each task from a single demonstration and from rehearsal on training scenes, and freezes what it learns as a Task Memory.
At deployment, it acts with the frozen memory on new scenes, anticipating how the task should unfold and recovering from failures.
After evaluation, a reflection agent reviews the records and writes guidance for the next round, realizing RSI over several iterations.
Data isolation is enforced by construction.
Every acting agent, whether in rehearsal or in deployment, observes only camera images with depth and proprioception, the same inputs available on a real robot.
Privileged simulator information serves only as supervision during learning, and a memory linter rejects coordinates, ground-truth terms, and seed identifiers.
This strict separation ensures that what is learned in simulation is directly deployable on a physical robot.
As shown in \figref{fig:infra}, the method, environment, embodiment, and foundation model are modular like Lego blocks.
Each runs as an independent service and can be replaced, making \method a reusable platform for agentic manipulation.

We conduct extensive experiments and present the main results on all 50 tasks of RoboTwin~2.0, leading to five findings.
1) One demonstration with rehearsal is sufficient.
With one clean demonstration per task, \method reaches 83.4\% mean success across the clean and randomized settings without gradient updates, whereas the strongest trained action policy, trained with 50 demonstrations, reaches 61.7\%.
2) What is learned in context generalizes across distributions.
Rehearsing only on clean scenes, \method maintains comparable success under domain randomization, whereas trained action policies retain at most 65.1\% of their clean success.
3) Reflection improves the frozen memory. Each round of reflection rescues more episodes than it breaks on difficult tasks.
4) The framework improves with the foundation model. Without any change to code, prompts, or protocol, GPT-6.1-Sol reaches 83.6/83.2\% on clean/randomized scenes, GPT-5.6-Luna reaches 79.8/76.0\%, and even DeepSeek-V4.1-Flash, a lightweight model, reaches 77.8/78.4\%, surpassing the strongest trained action policy.
5) Task Memories transfer across embodiments. With a lightweight adapter that rewrites an existing memory for a new robot, \method deploys on a physical dual-arm robot without retraining or re-rehearsing.
These findings jointly show that with a world in the loop, a robot can learn a manipulation task in its context as well as it learns one in its weights.

Our contributions are as follows.
\begin{itemize}[leftmargin=*, itemsep=2pt, topsep=2pt]
  \item We formulate WICL, a learning paradigm that extends in-context learning with feedback from a world.
  WICL decouples task skill from network weights and opens a path toward recursive self-improvement without retraining.
  We explain why the resulting knowledge transfers across worlds.
  \item We build a modular infrastructure for agent-centric manipulation with strict data isolation that ensures deployability on real hardware.
  We implement WICL on \method with a lifecycle covering skill building, task learning, deployment, and recursive self-improvement, all without gradient updates.
  \item We conduct extensive experiments with one demonstration per task and report results on RoboTwin~2.0, across distributions, over multiple rounds of self-improvement, with three foundation models, and on a physical dual-arm robot via cross-embodiment memory transfer, achieving state-of-the-art performance.
\end{itemize}
 \section{Related Work}
\label{sec:related}

\parahead{Action Policies for Manipulation}
The dominant approach to robot manipulation trains an action policy and stores skill in network weights.
Vision-language-action (VLA) models learn visuomotor skills end to end from demonstrations, typically after pretraining on large cross-embodiment datasets \citep{zitkovich2023rt2,kim2024openvla,black2024pi0,black2025pi05,bjorck2025gr00t,liu2024rdt,xie2026turbovla,openxembodiment2024,khazatsky2024droid,bu2025agibot}, with the VLA paradigm also advancing autonomous driving \citep{fu2025orion,fu2025minddrive,zhou2026qwendrive10initialstepvisionlanguage} and benefiting from efficient 3D representations \citep{liang2024pointmamba,liang2024pointgst}.
World action models (WAMs) generate future observations or latent plans and derive actions from them \citep{du2023unipi,cheang2024gr2,ye2026dreamzero,kim2026cosmospolicy,wu2026vega}, and driving world models jointly predict future scenes and 3D structure \citep{zhou2025hermes,zhou2026hermespp,liang2026UniFuture}.
Reinforcement learning optimizes a policy from simulator feedback \citep{andrychowicz2020inhand,zhao2020sim2real,li2026simplevlarl} or from real-robot trials \citep{kalashnikov2018qtopt,luo2024serl,luo2025hilserl}.
All three families have advanced rapidly, yet they share a structural limitation.
Because skill resides in weights, trained policies degrade under distribution shift \citep{chen2025robotwin,zhang2025vlaarena,gao2026ebench,zhou2025liberopro,fang2026towards}, and each new task or domain requires additional data and gradient updates.
WICL offers an alternative route.
It keeps the weights frozen and stores what it learns in context, so a new task requires rehearsal rather than retraining.

\parahead{In-Context Manipulation}
In robotics, in-context learning has mainly taken the form of imitation.
One-shot and few-shot methods interpret a demonstration given at test time \citep{duan2017oneshot,finn2017oneshotvisual,james2018tecnets,vosylius2025instant,oh2025rip,fu2025icrt}, while other methods pass keypoints or keyframes to a frozen model \citep{dipalo2024kat,yin2025roboprompt,palma2026bicicle} or retrieve demonstrations for a trained policy \citep{sridhar2025ricl,xu2026stellavla}.
Recent work has shown that in-context manipulation from a single demonstration is increasingly effective \citep{generalist2026gen15,skild2026s1,zhou2026zerowam}.
A parallel line uses frozen models as agents.
Training-free systems compose APIs, value maps, or keypoints \citep{liang2023codeaspolicies,huang2023voxposer,huang2024rekep,liu2024moka,fu2026capx}, while hierarchical systems place a language model above a trained policy \citep{ahn2022saycan,huang2022innermonologue,driess2023palme,shi2025hirobot,chen2026volo,liang2026cook}.
Embodied harnesses add tools, verification, and memory \citep{zhang2026harnessvla,liu2026guava,huang2026roboharness,gu2026harnesswam,wang2026thea}.
These methods place demonstrations, instructions, or tool descriptions in the context, but none distills experience from a world into a reusable memory.
Among these, HarnessVLA learns from interaction but depends on a trained policy for contact \citep{zhang2026harnessvla}, and FAEA reasons iteratively within an episode but reads privileged simulator state \citep{tsui2026faea}.
In contrast, \method removes the trained policy from the loop and enforces data isolation by construction.
WICL adds the missing feedback.
Where ICL gives the context a demonstration, WICL also gives it outcomes from a world, and the acting agent perceives the scene through cameras only.

\parahead{Self-Improving Agents and World Feedback}
Language model agents can improve from interaction while their weights stay frozen.
Reflexion stores reflections on failed trials, Voyager grows a library of code skills, and ExpeL extracts insights from past trajectories \citep{shinn2023reflexion,wang2023voyager,zhao2024expel}.
Later work rewrites the context itself \citep{wang2025awm,zhang2025ace,agrawal2026gepa}.
Coding agents solve real software tasks through the same loop \citep{yang2024sweagent,wang2024openhands}, though they learn within a single episode and do not distill experience across episodes.
These methods work in text and code.
WICL brings the same principle to physical manipulation, where the learned knowledge is semantic and transfers across worlds.
In-context reinforcement learning trains a model on prior interaction data so that it improves within its context at test time \citep{laskin2023ad,lee2023dpt}, but it requires this supervised pretraining.
In manipulation, self-improving systems collect practice data and retrain their policies \citep{bousmalis2023robocat,zhou2024soar,ghasemipour2025sief,torne2025casher}, and frozen language models have used simulator feedback to write reward functions that then train policies \citep{ma2024eureka}.
All of these use world feedback to update weights.
World models have been used to learn behaviors in imagination \citep{hafner2025dreamerv3,wu2022daydreamer,jang2025dreamgen}, to train policies with reinforcement learning \citep{zhu2026wmpo}, to evaluate existing policies \citep{guo2025ctrlworld,li2025worldeval}.
WICL uses the world differently.
It feeds outcomes back into the context with frozen weights, distills experience across episodes into a reusable memory, and extends the loop to recursive self-improvement without gradient updates.
 \section{World In-Context Learning}
\label{sec:wicl}
\begin{figure}[t]
  \centering
  \includegraphics[width=\textwidth]{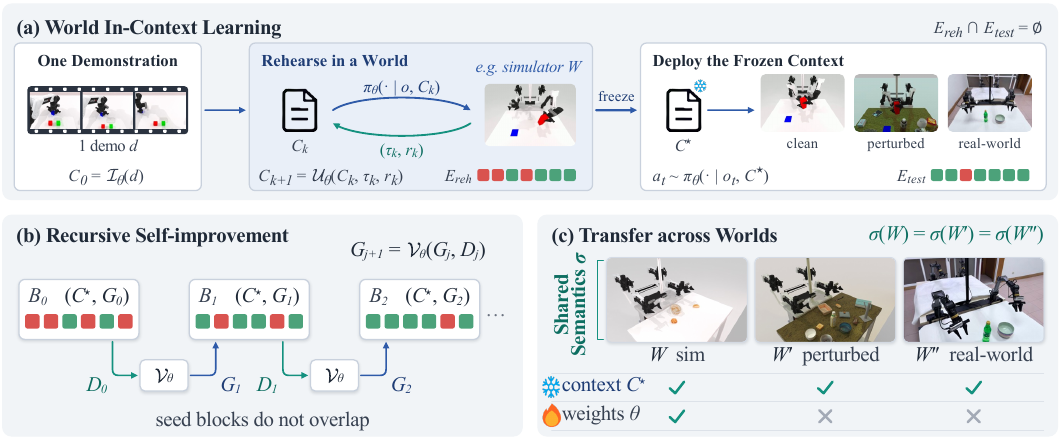}
  \caption{
  (a) The WICL pipeline. From one demonstration, the agent rehearses in a world, freezes the context, and deploys it on unseen instances. 
  (b) Recursive self-improvement over disjoint evaluation blocks with frozen weights. 
  (c) Context knowledge refers to shared semantics $\sigma$ and transfers across worlds, whereas trained weights do not.}
  \label{fig:wicl}
  \vspace{-0.4cm}
\end{figure}
 
\parahead{Formulation}
WICL extends in-context learning with feedback from a world, in the same way that reinforcement learning extends imitation learning.
Let $\pi_\theta$ be a foundation model with frozen weights $\theta$ that acts conditioned on an observation $o$ and a context $C$.
A task is specified by a demonstration $d$ and a set of instances, split into a rehearsal set $\mathcal{E}_{\mathrm{reh}}$ and an evaluation set $\mathcal{E}_{\mathrm{test}}$ with $\mathcal{E}_{\mathrm{reh}} \cap
\mathcal{E}_{\mathrm{test}} = \emptyset$. A world $W$ takes an instance $e$
and a policy, and returns a trajectory $\tau$ together with feedback $r$.
Learning starts from a context built from the demonstration and alternates between rehearsal and revision, as shown in \figref{fig:wicl}a,
\begin{equation}
  C_0 = \mathcal{I}_\theta(d), \qquad
  (\tau_k, r_k) \sim W\big(\pi_\theta(\cdot \mid o, C_k),\, e_k\big), \qquad
  C_{k+1} = \mathcal{U}_\theta(C_k, \tau_k, r_k),
  \label{eq:wicl-learn}
\end{equation}
where $e_k \in \mathcal{E}_{\mathrm{reh}}$ and $W$ supplies the observation $o$ at each step.
The same frozen model performs the initialization $\mathcal{I}_\theta$ and the update $\mathcal{U}_\theta$, so learning changes only the context.
Learning stops when a convergence rule is met, and the final context $C^\star$ is frozen.
The learner then acts on new instances with the frozen context and receives no feedback from the world,
\begin{equation}
  a_t \sim \pi_\theta(\cdot \mid o_t, C^\star), \qquad
  e \in \mathcal{E}_{\mathrm{test}}.
  \label{eq:wicl-eval}
\end{equation}
Since no evaluation instance is rehearsed, $C^\star$ is tested on generalization, not memorization.
In practice, $C$ is natural-language text that the model reads at every step.

\parahead{Transfer across Worlds}
The knowledge in $C^\star$ transfers across worlds because it is semantic, as illustrated in \figref{fig:wicl}c.
Let $\sigma(e)$ denote the semantic content of an instance, namely the objects present, their roles, the instruction, and the relative geometry on which the task depends.
The context $C^\star$ refers only to $\sigma$ and to quantities that are measured at execution time.
Suppose a second world $W'$ agrees with $W$ on the semantic content of its instances and on how behaviors change it, that is, $\sigma(W) = \sigma(W')$.
Then the knowledge rehearsed in $W$ still applies in $W'$, even if $W'$ differs in appearance, lighting, or physics parameters.
Weights trained in $W$ behave differently.
They absorb every regularity of $W$ that helps fit the training data, including regularities that $W'$ does not share.
This asymmetry is why WICL can learn in a cheap world and deploy in another, whether that is a randomized simulator or a real robot.

\parahead{Recursive Self-Improvement}
The learner can also improve from its own evaluations, as shown in \figref{fig:wicl}b.
Let $\mathcal{B}_0,
\mathcal{B}_1, \dots$ be disjoint blocks of evaluation instances and let $G_0
= \emptyset$.
After evaluating on $\mathcal{B}_j$ with context $(C^\star, G_j)$, the learner reflects on the resulting records $D_j$ and writes guidance for the next block,
\begin{equation}
  G_{j+1} = \mathcal{V}_\theta(G_j, D_j).
  \label{eq:wicl-rsi}
\end{equation}
The records $D_j$ contain only what a deployed system would observe, together with whether each episode succeeded.
They do not say why an episode failed, so the learner must diagnose the causes itself.
Each block is evaluated before the learner sees its outcomes, and the next block is always unseen.
This realizes recursive self-improvement with frozen weights.

\parahead{What a World Must Provide}
A world is useful for WICL if it meets three conditions.
It must be faithful to the task, so that behaviors change the semantic content $\sigma$ in the world as they would in the target environment.
It must be resettable, so that a behavior can be retried and compared.
And it must keep rehearsal instances apart from evaluation instances.
When no world is queried, WICL reduces to ICL, with $C^\star = C_0 = \mathcal{I}_\theta(d)$.
In this paper, the world is the RoboTwin~2.0 simulator \citep{chen2025robotwin}, and \cref{sec:method} describes our implementation.
 \section{\method}
\label{sec:method}
\begin{figure}[t]
  \centering
  \includegraphics[width=\textwidth]{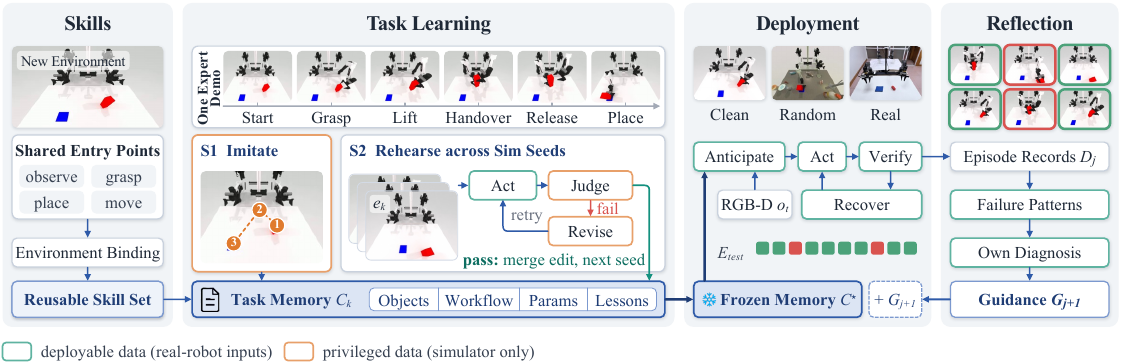}
  \caption{The four stages of \method and the data tier of each role.
  The agent builds reusable skills, learns a Task Memory from one demonstration through imitation (S1) and rehearsal (S2), deploys the frozen memory with anticipation and recovery, and improves through reflection over disjoint evaluation blocks.
  Green borders mark deployable data (real-robot inputs), and orange borders mark privileged data (simulator only).}
  \label{fig:overview}
  \vspace{-0.4cm}
\end{figure}
 
\method implements WICL for robot manipulation.
\figref{fig:overview} shows the pipeline together with the data tier of each role.
The world $W$ is a simulator, the demonstration $d$ is a single expert trajectory, and the learned context $C^\star$ is a Task Memory.
The agent builds its action interface (\cref{sec:perception}), learns a Task Memory from one demonstration (\cref{sec:learning}), and deploys it on unseen instances (\cref{sec:deployment}).
It can then improve the frozen memory through reflection (\cref{sec:rsi}).
Infrastructure and protocol parameters are given in the appendix.

\subsection{Perception and action}
\label{sec:perception}

The agent perceives and acts only through tools, and the same tools run in simulation and on a real robot.

\parahead{Geometric-first perception}
The agent observes RGB-D images from head and wrist cameras together with proprioception.
Object detection starts from the depth channel.
A height-map above the workspace produces connected-component proposals, each lifted to a 3D bounding volume with an attribute card that records its center, extent, and orientation.
Object identity is resolved by a three-way vote among Set-of-Mark annotation, a grounding model, and a VLM.
Once identified at the start of an episode, an object is tracked by its grasp relation to the end effector instead of re-detected after every action.
Part-level localization follows the same pipeline on a cropped region.

\parahead{Action interface with receipts}
The agent acts through a three-layer API.
Shared entry points for grasping, placing, corrective motions, reachability checks, and coordinated bimanual motion are environment-independent.
An environment binding discovered by automated probes supplies coordinate frames, joint limits, table height, and gripper characteristics.
Before learning, the agent builds environment-specific skills on these two layers.
It segments each demonstration into action primitives, identifies which segments the existing entries cannot explain, and hands each gap to a coding agent that writes a skill.
A driver validates each skill by running it on the demonstration scene and checking that it transfers to other layouts without encoding any absolute position.
The loop repeats until coverage stops improving, and the final skill set is frozen before task learning.

Every action returns a receipt reporting the commanded pose, the achieved pose, the measured gripper opening, and whether an object is held.
Receipts make the feedback $r$ in \eqnref{eq:wicl-learn} concrete.
Agent distinguishes a command that was executed from an effect that was achieved to decide its next step.
Receipts are derived from proprioception and the motion planner, not from simulator state, so they match on a real robot.

\subsection{Task learning from one demonstration}
\label{sec:learning}

The agent learns a Task Memory $C^\star$ from a single demonstration $d$ in two stages that together realize \eqnref{eq:wicl-learn}.
Both stages rehearse in the world, differing only in which instances the agent faces.
As shown in \figref{fig:overview}, border color marks the data tier of each role.

\parahead{Task Memory as semantic knowledge}
The Task Memory is a structured Markdown document that describes the roles of the objects and how to find them, the workflow of each stage, specifications for grasping and placing, learned parameters, and lessons from past attempts.
GraspSpec and PlaceSpec are relative specifications expressed in terms of perceived object parts, such as ``approach the handle from the left and grasp at its midpoint.'' They never contain positions in a particular scene.
This representation is the concrete form of the semantic knowledge in \cref{sec:wicl}.
The memory refers only to object roles and relative geometry and therefore applies to any scene that contains the same objects in reachable positions.
A memory linter enforces this property by rejecting coordinates, ground-truth terms, and instance identifiers.

\parahead{Imitation (S1)}
Imitation starts from the initialization $C_0 = \mathcal{I}_\theta(d)$.
A reader role studies the demonstration and writes an initial Task Memory.
An acting role then practices the task on the demonstration instance, and a reviewing role revises the memory after each attempt until the agent succeeds repeatedly.

\parahead{Rehearsal (S2)}
Rehearsal extends this to new training instances, realizing $C_{k+1} = \mathcal{U}_\theta(C_k, \tau_k, r_k)$.
The agent practices on one instance after another, and only its first attempt on each instance counts.
After a failure, the reviewing role receives a ground-truth diagnosis and revises a cached copy of the memory.
The revision is merged into the shared memory only if it makes that instance succeed.
Edits are typed and subject to a budget that decreases with experience, so the memory stabilizes as the agent practices.
Rehearsal stops when enough recent first attempts succeed, and the memory is frozen after passing the linter.

\parahead{Data isolation during learning}
As \figref{fig:overview} indicates, every acting agent observes only deployable data, the same inputs available on a real robot.
Privileged data enters only as supervision.
In S1, the reader uses it to understand the demonstration.
In S2, the reviewer receives a ground-truth diagnosis of each failure.
What this supervision teaches reaches the acting agent only through the Task Memory, and the linter ensures that the memory holds semantic knowledge, not simulator facts.
The actor therefore runs entirely on data that a real robot can provide and never accesses any simulator-specific privileged information.
This isolation ensures that the learned memory is not specific to any simulator.

\subsection{Deployment with anticipation and recovery}
\label{sec:deployment}

At deployment, the agent acts with the frozen memory $C^\star$ on instances it has never seen and receives no feedback $r$ from the world, as defined by \eqnref{eq:wicl-eval}.
It runs under the official protocol of the benchmark with unseen instructions.
As \figref{fig:overview} shows, the deployed actor uses only deployable data.

\parahead{Scene anticipation}
Before manipulating an object, the agent maps the demonstration's semantic blueprint onto the current scene.
It identifies which object plays each role in the memory, checks which arm can reach each grasp and placement, and verifies that the planned sequence is feasible.
This step may call the motion planner but never receives feedback about success and uses only deployable data.

\parahead{Failure recovery without reset}
A failed action does not end the episode.
The agent moves to a safe pose, observes the scene again, and continues from the current state without resetting the environment, bounded only by the official step limit.
The agent declares the task done only after its perception confirms that the scene matches the final state of the demonstration.
The simulator's success check is the final judge, so a wrong declaration never counts as a success.
On a real robot, the same procedure runs under human supervision.

\subsection{Recursive self-improvement}
\label{sec:rsi}

The reflection loop realizes the recursive self-improvement defined by \eqnref{eq:wicl-rsi}.
After an evaluation block, a reflection agent reads the episode records, which contain only what a deployed robot would observe together with whether each episode succeeded.
It identifies failure patterns and their causes and writes a guidance document for the next block.
Since the blocks never overlap, each round tests whether the guidance generalizes to unseen instances.
Repeating this loop over several rounds lets the agent improve without gradient updates or privileged data.
No instance is ever revisited, so the improvement measures generalization, not memorization.
 \section{Experiments}
\label{sec:experiments}
\definecolor{mainblock}{HTML}{F2F2F2}
\definecolor{mainours}{HTML}{F4F7FC}
\newcommand{\mainblock}[1]{\rowcolor{mainblock}\multicolumn{8}{l}{\textbf{#1}}}
\newcommand{\maingroup}[1]{\multicolumn{8}{l}{\textcolor{gray}{\textit{#1}}}}
\begin{table}[t]
  \tablestyle{3pt}{1.15}
  \caption{\textbf{Main results on RoboTwin~2.0.} Mean success rate (\%) over all 50 tasks.
  Retention is the randomized success divided by the clean success.
  Trained marks whether a method uses a trained action policy, and Agent marks whether it uses an agent.
  Trained action policies are trained jointly on all tasks with 50 clean demonstrations per task.
  $^\dagger$Randomized setting only, five seeds per task.
  $N$ is unreported fine-tuning demonstrations for LingBot-VLA~\citep{wu2026pragmatic}.}
  \label{tab:main}
  \begin{tabular}{lccc*{4}{x{41}}}
    \toprule
    \multirow{2}{*}{Method} & \multirow{2}{*}{\makecell[c]{Demos\\[-1.5pt]per task}} & \multirow{2}{*}{Trained} & \multirow{2}{*}{Agent}
      & \multicolumn{3}{c}{Success Rate (\%)} & \multirow{2}{*}{\makecell[c]{Retention\\[-1.5pt](\%)}} \\
    \cmidrule(lr){5-7}
    & & & & Clean & Rand. & Avg. & \\
    \midrule
    \mainblock{Trained action policies} \\
    \maingroup{Vision-language-action models (VLAs)} \\
    \quad StarVLA~\citep{community2026starvla} & 50 & \cmark & \xmark & 46.5 & 3.2 & 24.8 & 6.8 \\
    \quad GR00T-N1.7~\citep{bjorck2025gr00t} & 50 & \cmark & \xmark & 43.6 & 20.7 & 32.2 & 47.5 \\
    \quad GalaxeaVLA~\citep{galaxea2025} & 50 & \cmark & \xmark & 62.7 & 12.7 & 37.7 & 20.3 \\
    \quad Xiaomi-Robotics-0~\citep{cai2026xiaomi} & 50 & \cmark & \xmark & 62.9 & 18.2 & 40.6 & 28.9 \\
    \quad Abot-M0~\citep{yang2026abot} & 50 & \cmark & \xmark & 57.4 & 30.4 & 43.9 & 52.9 \\
    \quad X-VLA~\citep{zheng2025xvla} & 50 & \cmark & \xmark & 68.0 & 20.9 & 44.5 & 30.7 \\
    \quad Spatial Forcing~\citep{li2026spatial} & 50 & \cmark & \xmark & 77.2 & 26.7 & 52.0 & 34.6 \\
    \quad $\pi_{0.5}$~\citep{black2025pi05} & 50 & \cmark & \xmark & 70.7 & 46.0 & 58.4 & 65.1 \\
    \cmidrule(l){1-8}
    \maingroup{World action models (WAMs)} \\
    \quad AHA-WAM~\citep{cai2026ahawam} & 50 & \cmark & \xmark & 64.3 & 3.2 & 33.8 & 5.0 \\
    \quad FastWAM~\citep{yuan2026fastwam} & 50 & \cmark & \xmark & 77.8 & 1.9 & 39.9 & 2.4 \\
    \quad X-WAM~\citep{guo2026xwam} & 50 & \cmark & \xmark & 70.0 & 25.8 & 47.9 & 36.9 \\
    \quad LingBot-VA~\citep{lingbot-va2026} & 50 & \cmark & \xmark & 80.7 & 34.6 & 57.7 & 42.9 \\
    \quad 4D-WAM~\citep{yang20264dwaminfusingspatiotemporalawareness} & 50 & \cmark & \xmark & \underline{81.5} & 41.8 & 61.7 & 51.3 \\
    \midrule
    \mainblock{Agents with trained policies} \\
    \quad HarnessVLA (frozen base)~\citep{zhang2026harnessvla}$^\dagger$ & $N$ & \cmark & \xmark & -- & 50.4 & -- & -- \\
    \quad HarnessVLA (Codex)~\citep{zhang2026harnessvla}$^\dagger$ & $N$\,+\,1 & \cmark & \cmark & -- & 58.0 & -- & -- \\
    \quad HarnessVLA (Claude Code)~\citep{zhang2026harnessvla}$^\dagger$ & $N$\,+\,1 & \cmark & \cmark & -- & 58.4 & -- & -- \\
    \midrule
    \mainblock{Action-policy-free agents} \\
    \rowcolor{mainours} \quad \method (DeepSeek-V4.1-Flash) & 1 & \xmark & \cmark & 77.8 & \underline{78.4} & \underline{78.1} & 100.8 \\
    \rowcolor{mainours} \quad \method (GPT-5.6-Luna) & 1 & \xmark & \cmark & 79.8 & 76.0 & 77.9 & 95.2 \\
    \rowcolor{mainours} \quad \method (GPT-6.1-Sol) & 1 & \xmark & \cmark & \textbf{83.6} & \textbf{83.2} & \textbf{83.4} & 99.5 \\
    \bottomrule
  \end{tabular}
\end{table}
 
We conduct extensive experiments in simulation and on real hardware, organizing the evaluation around six questions.
Can one demonstration with rehearsal compete with policies trained on many demonstrations (\cref{sec:exp_main})?
Why does success hold under domain randomization (\cref{sec:exp_random})?
Where do the gains come from (\cref{sec:exp_ablation})?
Can the agent improve from its own evaluations (\cref{sec:exp_rsi})?
What does it cost (\cref{sec:exp_models})?
Does the learned knowledge transfer across embodiments to a real robot (\cref{sec:exp_real})?

\subsection{Experimental Setup}
\label{sec:exp_setup}

\parahead{Benchmark and protocol}
RoboTwin~2.0 \citep{chen2025robotwin} defines 50 dual-arm tasks in a clean setting (fixed background) and a randomized setting (randomized textures, lighting, table height, and distractors).
The agent first builds its skill set across all 50 tasks, then learns each task from one automatically selected clean demonstration, with rehearsal on clean scenes only.
The frozen Task Memory is evaluated in both settings with the official server, unseen instructions, and the official step limit.
Each task is evaluated on 10 episodes per setting (1,000 episodes total), as each episode requires dozens of model calls.
We report success rate averaged over all tasks, the mean of the two settings, and retention (randomized divided by clean).

\parahead{Baselines}
Trained action policies (VLAs and WAMs) are trained on all 50 tasks with 50 clean demonstrations per task and evaluated on 100 episodes.
The agent-with-policy baseline is HarnessVLA \citep{zhang2026harnessvla}, which orchestrates a fine-tuned VLA with a coding agent.
All baseline results come from the official leaderboard \citep{chen2025robotwin} or published reports, excluding policies trained on randomized data.

\parahead{Foundation models and model service}
We run \method with GPT-6.1-Sol, GPT-5.6-Luna, and DeepSeek-V4.1-Flash with identical code, prompts, and protocol.
API service quality varies over time and can shift success rates by several points between sessions.
For the ablation study (\cref{sec:exp_ablation}), each group of variants and its control share one fixed endpoint and the same seeds, so within-group comparisons are paired, though absolute levels may differ across groups.
Further settings are in \cref{sec:app_method}.

\subsection{Main Results}
\label{sec:exp_main}

\tabref{tab:main} compares \method with trained action policies and agents with trained policies on all 50 tasks.

\parahead{One demonstration with rehearsal is sufficient}
With one demonstration per task and no gradient update, \method (GPT-6.1-Sol) reaches a mean of 83.4\% (83.6 clean, 83.2 randomized), with a 95\% Wilson interval of 81.0 to 85.6.
The strongest trained policy, 4D-WAM trained on 50 demonstrations per task, reaches 61.7\% (81.5 clean, 41.8 randomized).
The two are comparable in the clean setting, yet the mean differs by 21.7 points because trained policies lose most of their success under randomization.
These gains come from a single demonstration and rehearsal in a simulator, without any training data, gradient computation, or access to the randomized distribution.

\parahead{What is learned in context holds across distributions}
The three foundation models retain 95.2\% to 100.8\% of their clean success under randomization, whereas trained policies retain at most 65.1\%.
The Task Memory stores semantic knowledge that does not depend on visual appearance, so it applies equally to both settings.
Even though the agent rehearses only on clean scenes and never observes any randomized instance, \method exceeds HarnessVLA (58.4\% randomized), which fine-tunes its base policy on RoboTwin data, by 24.8 points.
\cref{sec:exp_random} analyzes where the remaining failures come from.

\parahead{The advantage does not depend on the strongest foundation model}
DeepSeek-V4.1-Flash and GPT-5.6-Luna reach means of 78.1\% and 77.9\%, both above every trained policy.
GPT-6.1-Sol leads them by 5.3 and 5.5 points, and a paired bootstrap over tasks keeps both gaps above zero (95\% intervals 0.7 to 10.0 and 0.8 to 10.2).
The two weaker models differ by only 0.2 points.
All three use the same code, prompts, and protocol, suggesting that performance scales with model capability.
Since the knowledge lives in the context and not in the weights, any foundation model with sufficient capability can serve as the base.
\cref{sec:exp_models} reports the cost of each model.

\subsection{Why Does Success Hold under Randomization?}
\label{sec:exp_random}
\begin{figure}[t]
  \centering
  \includegraphics[width=\textwidth]{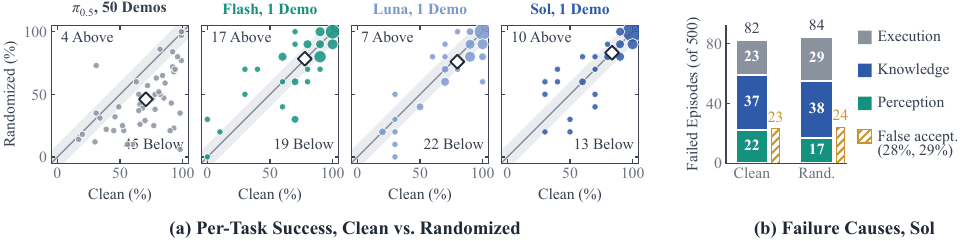}
  \caption{\textbf{Success under randomization.} (a) Per-task success. Dot area counts tasks at one point,
  the band marks a change within 10 points, and the diamond marks the mean. (b) Failed episodes of GPT-6.1-Sol by physical root cause.
  Hatched bars count false acceptances, which cut across the causes.}
  \label{fig:random}
  \vspace{-0.2cm}
\end{figure}
 \begin{figure}[t]
  \centering
  \includegraphics[width=\textwidth]{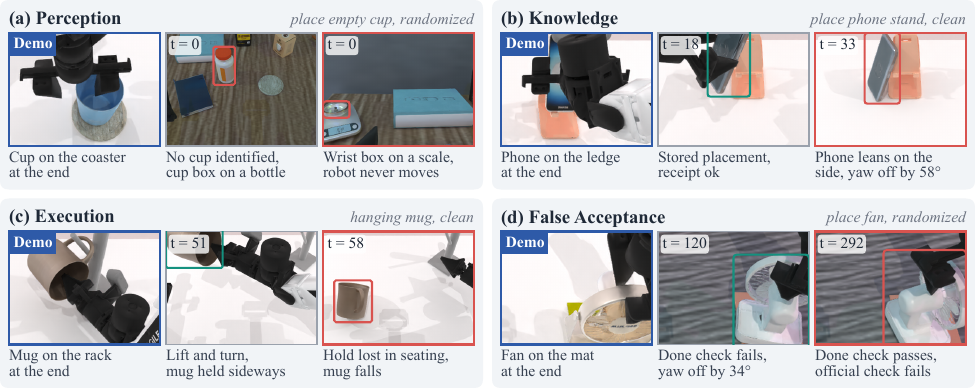}
  \caption{\textbf{Failure cases.} One failed evaluation episode of GPT-6.1-Sol for each cause and for false acceptance, with $t$ the simulator step.
  Blue marks the demonstration and red the failure.
  (a) The cup is never found. (b) The stored placement leaves the phone leaning. (c) The mug drops while seated. (d) The agent's completion check accepts a fan that the official check rejects.}
  \label{fig:failure}
  \vspace{-0.2cm}
\end{figure}
 
\begin{wrapfigure}{r}{0.52\linewidth}
  \vspace{-0.4cm}
  \centering
  \includegraphics[width=\linewidth]{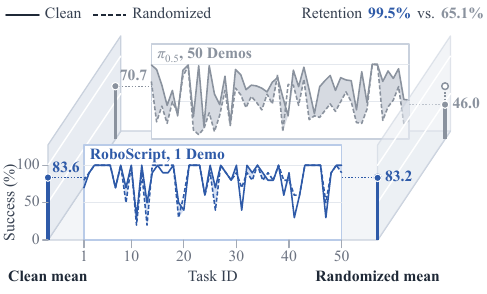}
  \caption{\textbf{Performance retention under randomization.} Per-task success of \method (front) and $\pi_{0.5}$ (back), with the drop from clean to randomized shaded.
}
  \label{fig:retention}
  \vspace{-0.4cm}
\end{wrapfigure}

\method rehearses only on clean scenes, yet retains over 95\% of its success under randomization (\tabref{tab:main} and \figref{fig:retention}).
We verify this with per-task analysis and failure diagnosis.

 \parahead{Per-task success is stable across settings}
\figref{fig:random}a plots clean versus randomized success for each task.
For $\pi_{0.5}$, most tasks fall below the diagonal, and every task category loses 18.7 to 31.8 points under randomization.
The three foundation models instead scatter on both sides of the diagonal.
Averaged within each category, their success changes by at most 8.0 points between settings, and for GPT-6.1-Sol the change ranges from $-$7.0 to $+$6.4.

\parahead{Knowledge failures do not increase under randomization}
We label each failed GPT-6.1-Sol episode with a root cause (perception, knowledge, or execution) from its event log and check for false acceptances (\cref{sec:app_additional_results}).
\figref{fig:random}b shows the distribution over 166 failed episodes (82 clean, 84 randomized).
Knowledge failures account for 37 and 38, nearly unchanged between settings.
Perception failures decrease slightly (22 to 17) and execution failures increase (23 to 29), consistent with randomized table heights making motions harder without affecting what the agent knows.
False acceptance rates are also stable (28\% and 29\%).
\figref{fig:failure} shows one representative failure for each cause.
The failures under randomization are physical, not epistemic, consistent with the semantic nature of the learned knowledge.

\subsection{Where Do the Gains Come From?}
\label{sec:exp_ablation}

\definecolor{ablblock}{HTML}{F2F2F2}
\definecolor{ablfull}{HTML}{F4F7FC}
\newcommand{\ablblock}[1]{\rowcolor{ablblock}\multicolumn{4}{l}{\textbf{#1}}}
\begin{table}[!t]
  \noindent\begin{minipage}{\linewidth}
  \centering
  \includegraphics[width=\linewidth]{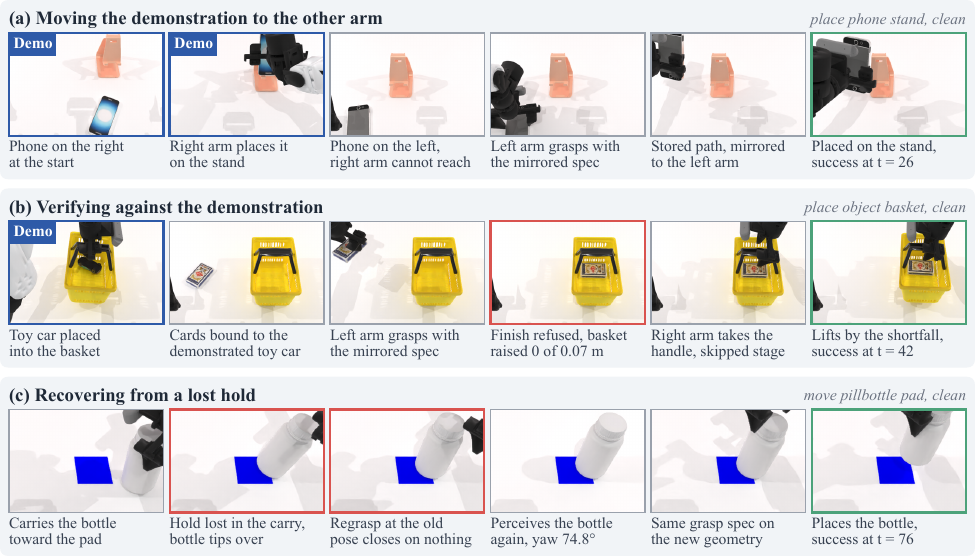}
  \captionof{figure}{\textbf{Deployment with a frozen Task Memory.} Evaluation episodes of GPT-6.1-Sol, with $t$ the simulator step.
  Blue marks the demonstration, red an error, and green success.
  (a) The agent mirrors the demonstration to the other arm. (b) A refused finish leads it to the skipped stage. (c) It grasps the dropped bottle again.}
  \label{fig:qual}
\end{minipage}
 \tablestyle{6pt}{1.15}
  \caption{\textbf{Ablations.} Success rate (\%) of GPT-6.1-Sol on 10 tasks, with 10 episodes per task in each setting.
  Each variant is compared with the shaded full method above it on the same seeds.
  The second group builds the memory without interacting with the world, and the third group learns every memory from scratch.}
  \label{tab:ablation}
  \begin{tabular}{l*{3}{x{40}}}
    \toprule
    \multirow{2}{*}{Variant} & \multicolumn{3}{c}{Success Rate (\%)} \\
    \cmidrule(lr){2-4}
    & Clean & Rand. & Avg. \\
    \midrule
    \ablblock{Deploying a frozen memory} \\
    \rowcolor{ablfull} \method & 77.0 & 80.0 & 78.5 \\
    \quad w/o arm check & 78.0 & 72.0 & 75.0 \\
    \quad w/o completion check & 72.0 & 82.0 & 77.0 \\
    \quad w/o recovery & 68.0 & 70.0 & 69.0 \\
    \midrule
    \ablblock{Without the world} \\
    \quad Demonstration written to memory & 69.0 & 69.0 & 69.0 \\
    \quad Demonstration in deployment context & 69.0 & 66.0 & 67.5 \\
    \quad Instruction only & 68.0 & 68.0 & 68.0 \\
    \midrule
    \ablblock{Learning} \\
    \rowcolor{ablfull} \method (relearned) & 87.0 & 85.0 & 86.0 \\
    \quad w/o demonstration & 58.0 & 55.0 & 56.5 \\
    \quad w/o privileged supervision & 87.0 & 86.0 & 86.5 \\
    \bottomrule
  \end{tabular}
  \vspace{-0.4cm}
\end{table}
 
\begin{figure}[t]
  \centering
  \includegraphics[width=\textwidth]{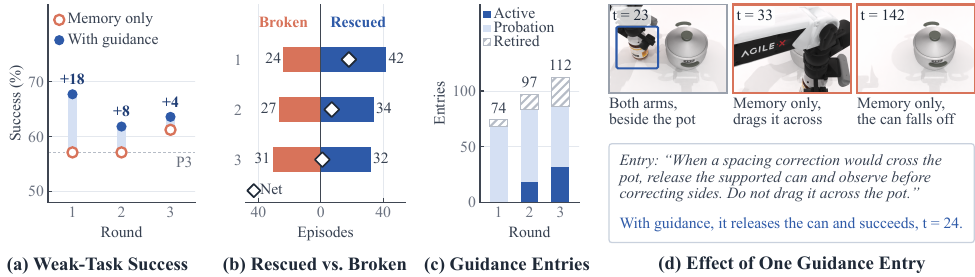}
  \caption{\textbf{Recursive self-improvement.} GPT-6.1-Sol over three rounds on new clean seeds shared by both arms.
  (a) Success on the 17 weak tasks, with P3 dashed. (b) Episodes of all 50 tasks that guidance rescues or breaks.
  (c) Guidance entries by status. (d) An entry added in round 3 changes the outcome on one seed.}
  \label{fig:rsi}
  \vspace{-0.2cm}
\end{figure}
 
\tabref{tab:ablation} removes one component at a time.
Deployment ablations use the frozen memory from the main run, the second group builds a memory without interacting with the world, and learning ablations relearn every memory from scratch.
All groups use the same 10 tasks covering all five categories, with 10 episodes per task in each setting on shared seeds, so all comparisons are paired and we report exact McNemar $p$ values.

\parahead{The demonstration and the world are both necessary}
The second group of \tabref{tab:ablation} separates the two inputs of WICL.
Writing the memory from the demonstration alone, without imitation or rehearsal in the world, gives 69.0\%, against 78.5\% when the same demonstration is followed by interaction with the world ($p=0.003$).
Without the world, the demonstration brings no benefit.
Showing it to the agent at deployment gives 67.5\%, and a memory that holds only the instruction gives 68.0\%, so the three variants differ by at most 1.5 points (pairwise $p \geq 0.76$).
The world in turn needs the demonstration.
Rehearsal from the instruction alone reaches 56.5\%, which is 11.5 points below the same instruction-only memory deployed without rehearsal ($p=0.004$).

\parahead{The demonstration anchors learning}
Without a demonstration, the agent begins rehearsal from an empty memory and the instruction alone, and the mean drops from 86.0\% to 56.5\%, the largest gap in the study.
Rehearsal from trial outcomes alone still solves three tasks entirely (\texttt{turn\_switch}, \texttt{click\_alarmclock}, and \texttt{move\_stapler\_pad}), but two others fail to converge within the budget and score 0.
The demonstration gives the agent a working starting point, without which convergence is unreliable.
This variant also removes the demonstration-anchored completion check, so the drop reflects both losses.

\parahead{Rehearsal corrects what the demonstration does not cover}
Rehearsal tests the memory on new scenes before deployment and corrects the specifications that imitation leaves wrong, such as grasp and placement parameters.
The 10 tasks of \tabref{tab:ablation} exclude the tasks that are especially costly and hard to learn, and the effect of rehearsal is clearest on those tasks.
On 11 such tasks, deploying the memory from imitation without rehearsal gives 46.1\%, against 54.5\% for the full method.
The gain is 8.4 points on average and reaches 13.3 points under randomization ($p=0.034$), where scenes differ most from the single demonstration.
The rehearsal protocol applies strict edit budgets and convergence rules, so it changes the shared memory only where repeated trials show an error.

\parahead{Anticipation and recovery at deployment}
\figref{fig:qual} illustrates the three deployment components that we ablate.
Removing recovery, so that an episode ends at its first failed action, produces the largest drop in this group (78.5\% to 69.0\%).
Removing the arm check, so that the agent always uses the arm from the demonstration instead of choosing per scene, reduces the mean to 75.0\% (78 clean, 72 randomized).
In the full method the agent switches arms in about half of successful grasps and achieves comparable success (93\% and 95\%), confirming that switching does not hurt.
Removing the completion check, so that the agent no longer verifies the final state against the demonstration, gives 77.0\% (clean $-$5, randomized $+$2). The clean loss comes from episodes that skip a manipulation stage.

\parahead{Privileged supervision is not the source of performance}
Learning has two privileged inputs.
The demonstration stage observes ground-truth object state, and the reviewing role in rehearsal reads a ground-truth diagnosis of each failed attempt.
We remove both at once and also restrict every role of the demonstration stage to deployable observations.
Only the official success signal that each task records and the keypoints from the robot trajectory of the demonstration remain.
This privilege-free variant reaches 86.5\%, against 86.0\% for the full method.
Since the acting agent also uses only deployable observations during rehearsal and evaluation, the learned Task Memory does not depend on privileged information.

\subsection{Recursive Self-Improvement}
\label{sec:exp_rsi}

We evaluate the RSI loop (\eqnref{eq:wicl-rsi}) on the frozen memories of GPT-6.1-Sol over three rounds.
Each round draws a new set of clean seeds with 10 episodes per task and evaluates the memory both with and without guidance on the same seeds, so the paired gap isolates the effect of reflection.
The reflection agent reads only deployable observations and whether each episode succeeded.

On the 17 weak tasks (clean success $\leq$8 of 10 before reflection), guidance rescues more episodes than it breaks in every round, with net gains of $+$18, $+$8, and $+$4 episodes (\figref{fig:rsi}a).
On all 50 tasks the net gains are $+$18, $+$7, and $+$1 (\figref{fig:rsi}b).
The gains concentrate on harder tasks where the reflection agent identifies failure patterns that the original memory did not address.
The number of guidance entries grows from 74 to 97 to 112 (\figref{fig:rsi}c).
\figref{fig:rsi}d shows one episode in which a round 3 entry changes the action the agent takes.
This confirms that frozen weights do not prevent improvement when the knowledge is stored in readable context.

\subsection{Cost of Learning and Deployment}
\label{sec:exp_models}
\begin{table}[t]
  \tablestyle{3pt}{1.15}
  \caption{\textbf{Foundation models and cost.} DS is DeepSeek, and Avg.\ is the success rate (\%) of \tabref{tab:main}.
  Exits counts the tasks that pass the S1 exit rule, and Instances is the mean number of S2 instances per task.
  Evaluation columns are means per episode, with tokens in millions. Cost covers learning and evaluation at list API price.}
  \label{tab:models}
  \begin{tabular}{l*{8}{x{42}}}
    \toprule
    \multirow{2}{*}{Model} & \multirow{2}{*}{Avg.} & \multicolumn{2}{c}{S1} & S2 & \multicolumn{3}{c}{Evaluation episode} & Cost \\
    \cmidrule(lr){3-4} \cmidrule(lr){5-5} \cmidrule(lr){6-8}
    & & Exits & Attempts & Instances & Calls & Tokens & Steps & (USD) \\
    \midrule
    DS-V4.1-Flash & 78.1 & 50/50 & 301 & 11.1 & 53.1 & 3.89 & 183 & 739 \\
    GPT-5.6-Luna & 77.9 & 47/50 & 518 & 14.5 & 62.2 & 2.61 & 190 & 308 \\
    GPT-6.1-Sol & 83.4 & 49/50 & 329 & 12.5 & 54.7 & 2.33 & 149 & 1,911 \\
    \bottomrule
  \end{tabular}
  \vspace{-0.4cm}
\end{table}
 
\tabref{tab:models} reports the success of each foundation model together with the cost of learning and deployment.

\parahead{Learning and deployment}
The demonstration stage takes 301 to 518 attempts across the three models, and rehearsal averages 11.1 to 14.5 instances per task.
GPT-6.1-Sol passes the demonstration exit rule on 49 of 50 tasks, DeepSeek-V4.1-Flash on all 50, and GPT-5.6-Luna on 47.
Tasks that do not pass are frozen with the best available memory (\cref{sec:app_additional_results}).
At deployment, an evaluation episode with GPT-6.1-Sol takes on average 54.7 model calls, uses 2.33 million tokens, and runs for 149 simulation steps.

\begin{figure}[t]
  \centering
  \includegraphics[width=\textwidth]{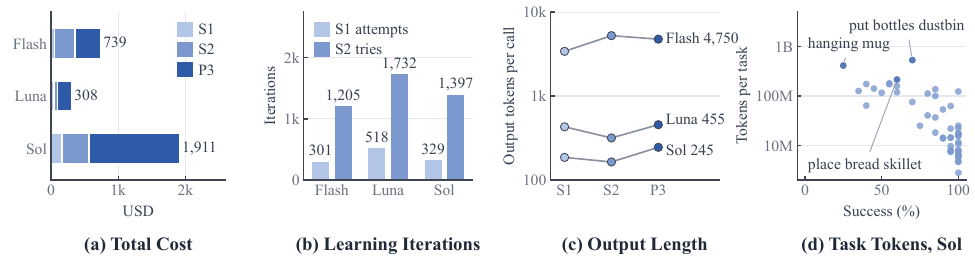}
  \caption{\textbf{What the cost consists of.} (a) Cost at list API price by stage, mostly from evaluation.
  (b) Total S1 attempts and S2 tries. (c) Output tokens per call of the actor. DeepSeek-V4.1-Flash writes about 19 times more than GPT-6.1-Sol in evaluation.
  (d) Per-task tokens and success of GPT-6.1-Sol. Less successful tasks take more tokens.}
  \label{fig:app_cost}
\end{figure}

\parahead{Cost}
At list API prices, learning all 50 tasks and evaluating 1,000 episodes costs 308 USD with GPT-5.6-Luna and 1,911 USD with GPT-6.1-Sol.
Evaluation accounts for the majority of the cost (\cref{fig:app_cost}).
DeepSeek-V4.1-Flash generates about 19 times more output tokens per call than GPT-6.1-Sol, so despite a lower per-token price its total reaches 739 USD.
Once frozen, the Task Memory can be reused across episodes without additional learning cost.

\subsection{Real-Robot Deployment}
\label{sec:exp_real}
\begin{wrapfigure}[11]{r}{0.25\linewidth}
  \vspace{-0.4cm}
  \centering
  \includegraphics[width=\linewidth]{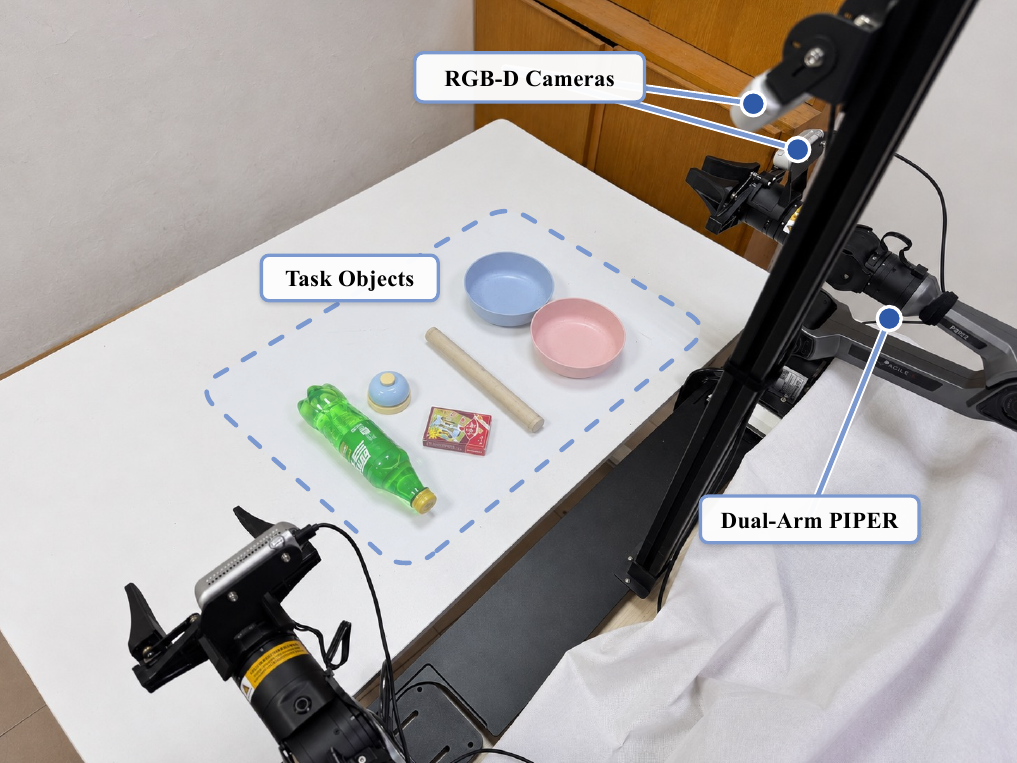}
  \caption{Real-robot setup with a dual-arm PiPER and RGB-D cameras.}
  \label{fig:real_setup}
  \vspace{-0.4cm}
\end{wrapfigure}
 \begin{figure*}[t]
  \centering
  \includegraphics[width=\textwidth]{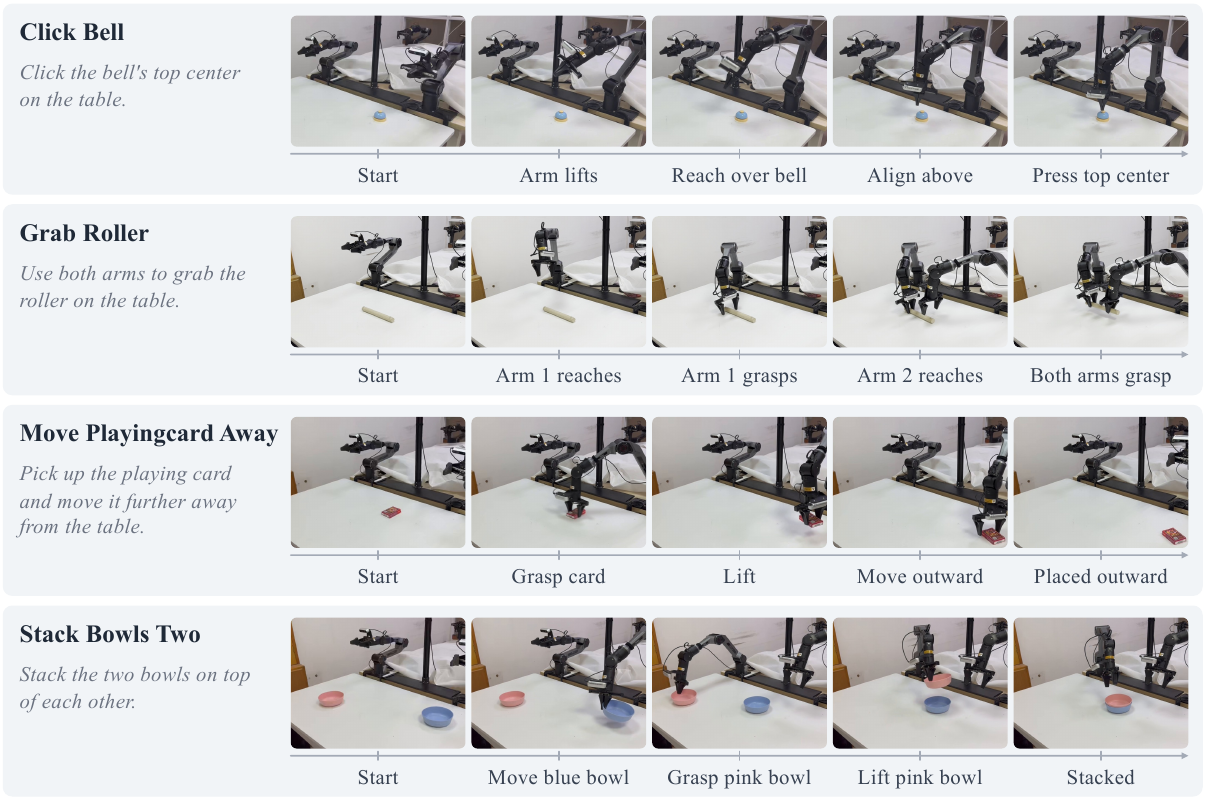}
  \caption{Execution frames for four real-robot tasks.
  Each row shows key frames from a successful episode.
  The Task Memory was rehearsed on a simulated ALOHA and transferred to the PiPER through a lightweight adapter without re-rehearsing.}
  \label{fig:real_rollout}
  \vspace{-0.3cm}
\end{figure*}
 
We deploy \method on a physical dual-arm PiPER robot (\figref{fig:real_setup}) to test whether Task Memories learned in simulation transfer to a different embodiment.
The PiPER has two 6-DoF arms with parallel-jaw grippers and a different kinematic structure than simulated ALOHA used during rehearsal.

\parahead{Setup}
We take the frozen Task Memories from the main experiment (GPT-6.1-Sol) and apply a lightweight adapter that remaps the embodiment-specific parameters for the PiPER robot.
The adapter maps coordinate conventions, arm kinematics, and gripper parameters while leaving the task workflow, object roles, and relative specifications unchanged.
The original demonstrations and rehearsal are not repeated.
We plan the full action sequence from the memory offline so the robot executes each step without online replanning, and test four tasks with 10 episodes each.

\parahead{Results}
\figref{fig:real_rollout} shows execution frames for the four tasks.
\method succeeds on 10 of 10, 7 of 10, 7 of 10, and 6 of 10 episodes on Click Bell, Grab Roller, Move Playingcard Away, and Stack Bowls Two, respectively.
These tasks cover pressing, bimanual grasping, pick-and-place, and stacking.
Failures are mainly caused by gripper misalignment and depth estimation errors on real sensors.
The memories were rehearsed entirely in simulation on a different robot, yet they produce working behavior on real hardware through parameter remapping alone.
This validates WICL's prediction that semantic knowledge transfers across worlds.
 \section{Conclusion}

We have introduced World In-Context Learning (WICL), a paradigm that extends in-context learning with feedback from a world, and implemented it in \method, a modular infrastructure for agentic manipulation with strict data isolation.
On all 50 tasks of RoboTwin~2.0, the agent builds its own skills, learns each task from one demonstration and rehearsal in a simulator, and reaches 83.4\% mean success across clean and randomized settings without gradient updates, exceeding every trained action policy.
The learned knowledge is semantic, so it transfers across visual distributions, works with three different foundation models, and improves through reflection over successive rounds.
A lightweight adapter transfers the same memories to a physical PiPER robot with a different kinematic structure, producing working behavior without re-recording any demonstration or repeating any rehearsal.
These findings suggest that task skill need not reside in network weights, and that with a world in the loop, a robot can learn a manipulation task in its context as well as it learns one in its weights.
 
\clearpage
\beginappendix

We first extend the discussion of WICL and its relation to other paradigms (\cref{sec:app_wicl}), then describe the system architecture (\cref{sec:app_architecture}) and the method details including skills, perception, learning, and data isolation (\cref{sec:app_method}), present additional quantitative (\cref{sec:app_additional_results}) and qualitative (\cref{sec:app_qualitative}) results, and finally discuss limitations and future work (\cref{sec:app_limitations}).

\noindent\hspace{1.5em}
\renewcommand{\arraystretch}{1.3}
\begin{tabular}{@{}ll@{}}
\cref{sec:app_wicl} & Extended Discussion about WICL \\
\cref{sec:app_architecture} & System Architecture \\
\cref{sec:app_method} & Method Details \\
\cref{sec:app_additional_results} & Additional Quantitative Results \\
\cref{sec:app_qualitative} & Additional Qualitative Results \\
\cref{sec:app_limitations} & Limitations and Future Work \\
\end{tabular}
\vspace{0.3cm}

\section{Extended Discussion about WICL}
\label{sec:app_wicl}

We discuss how WICL relates to established learning paradigms, what distinguishes it from prior work, and why it has become feasible for physical manipulation.

\parahead{Relation to Other Learning Paradigms}
Skill acquisition differs along two dimensions. The first is the source of learning, either demonstrations alone or feedback from a world. The second is where the learned skill is stored, either in network weights or in the context of a frozen model.
Imitation learning writes demonstrations into weights. Reinforcement learning adds a world whose feedback is also written into weights. In-context learning stores demonstrations in context. WICL adds a world whose feedback is also stored in context (\figref{fig:teaser}a).
The world provides a feedback signal that demonstrations alone cannot offer, because a demonstration is recorded in a different scene and cannot tell the learner what will happen in the scene in front of it. Writing this feedback into context instead of weights avoids the distribution coupling that limits weight-based methods.

\parahead{Distinction from In-Context RL and Experiential Learning}
In-context reinforcement learning \citep{laskin2023ad,lee2023dpt} also improves within the context, but its weights are first trained on learning histories from the target domain, so the model itself is specialized.
WICL uses a general foundation model without such training.
Experiential learning in LLM agents \citep{shinn2023reflexion,wang2023voyager,zhao2024expel} is the closest relative.
Some of these methods improve through repeated trials on the same instance \citep{shinn2023reflexion}, which conflates memorization with generalization.
The split between $\mathcal{E}_{\mathrm{reh}}$ and $\mathcal{E}_{\mathrm{test}}$ in WICL rules this out and ensures that every reported success measures generalization.

\parahead{WICL in Physical Manipulation}
Applying WICL to physical manipulation introduces requirements that do not arise in software or text domains.
The context must be grounded in camera observations of a scene that changes as the robot acts.
The feedback from the world includes physical outcomes such as whether an object was grasped, not only textual signals.
The context must therefore record task-relevant semantics from visual observations while remaining independent of the visual appearance of any particular world.
In our implementation, a linter enforces this by rejecting coordinates, ground-truth terms, and instance identifiers.

\parahead{Feasibility Conditions}
Simulators and robot APIs have been available for years \citep{liang2023codeaspolicies}, but WICL also requires a model that can use tools over a long context, read feedback, and accumulate experience.
The progress of coding agents on real software repositories shows that foundation models have crossed this threshold \citep{yang2024sweagent,wang2024openhands}.
Their terminal is their world, and a simulator can now serve the same role for a robot.
The remaining requirement is a world that is faithful to the task, resettable, and keeps rehearsal apart from evaluation instances.

\section{System Architecture}
\label{sec:app_architecture}

\begin{figure}[t]
  \centering
  \includegraphics[width=\textwidth]{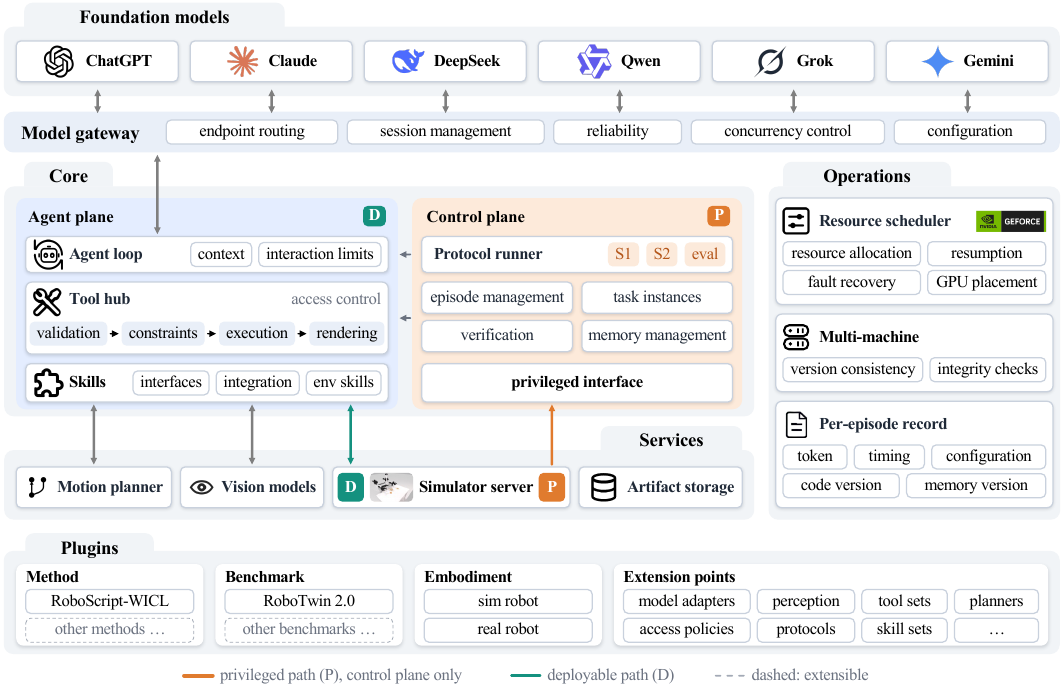}
  \caption{Detailed architecture of \method, expanding \figref{fig:infra}. The model gateway handles endpoint routing, session management, and reliability. The core separates an agent plane (agent loop, tool hub, skills) from a control plane (protocol runner, memory management, verification). Services, operations, and plugins are independently replaceable.}
  \label{fig:infra-detail}
\end{figure}
 
\parahead{Two planes}
As shown in \figref{fig:infra-detail}, \method separates a control plane from an agent plane.
The control plane runs the protocols of the lifecycle.
It opens and closes episodes, selects instances, judges success, and stores and freezes memories.
It is the only part of the system that can reach privileged simulator state.
The agent plane contains the foundation model, the tool hub, and the tools.
The simulator, the motion planner, and the vision models run as separate services.
Every message between processes is a typed and versioned contract, and large arrays are passed by reference.
Each component can therefore be developed, tested, and replaced on its own.

\parahead{Model gateway}
Every model call, whether from the acting agent, a reviewing role, or perception, passes through one gateway.
The gateway exposes a single interface over several provider endpoints.
It keeps a session on the endpoint that served it, moves to another endpoint when one fails, and adapts the number of concurrent requests to what each endpoint sustains.
Changing the foundation model of a run is a change of configuration, and the same setting applies to every role and to perception.

\parahead{Tool hub}
The hub sits between the model and the tools and applies the rules of each stage.
When an episode opens, the hub fixes the tools that each role may see from the stage, the role, and the data tier.
Every call then passes through one ordered pipeline that validates it, checks the budgets and the stage rules, executes it, and renders the result.
Evaluation stages admit tools of the deployable tier only, and the hub refuses any privileged call made in them, including a call that one tool makes on behalf of another.

\parahead{Scheduling and reproducibility}
For each task, a job graph runs imitation, rehearsal, and deployment in order.
A scheduler places jobs on GPUs according to their measured memory, starts the services of a GPU when a job first needs them, and resumes interrupted jobs from their records.
Infrastructure failures, such as a lost connection to a model endpoint, are retried and are not counted as task failures.
Across machines, tasks are assigned to workers that run the same code version and configuration, and their results are copied back and verified with checksums.
A run does not start from an unversioned or modified code tree.
Every episode records its code version, configuration, memory version, token usage, and duration, and the event log of an episode is sufficient to reconstruct everything the model saw.

\parahead{Extensibility}
All major components, including benchmarks, embodiments, perception backends, motion planners, model clients, and methods, are registered through plugin interfaces with conformance tests.
The method in this paper is one such plugin built on these mechanisms.

\section{Method Details}
\label{sec:app_method}

\subsection{Skills and Specifications}
\label{sec:app_skills}

\parahead{Shared and environment-specific skills}
Skills are divided by whether they depend on the environment.
Shared entry points cover common operations such as observing, grasping, placing, corrective motions, reachability checks, arm selection, and coordinated bimanual motion.
These entry points hold no convention of any environment.
Environment conventions, including coordinate frames, end-effector-to-fingertip offsets, gripper normalization, the collision world, and arm names, enter the shared entry points as data through a binding.
An operation that holds only in a particular environment is an environment-specific skill, and it becomes visible only when that environment is loaded.

\parahead{Specifications instead of poses}
Grasps and placements are described by specifications instead of poses.
A grasp specification names the part to hold and the approach relative to that part.
A placement specification names the target and the offset and orientation relative to it.
The frames of these specifications are perceived objects, their parts, landmarks, or the robot base, and no specification stores the absolute position of an object.
At execution time, the skill solves the specification on the perceived geometry of the current scene, mirrors it when the other arm acts, and keeps the first solution that the motion planner can reach.
The parameters of the specifications belong to the Task Memory and are learned during rehearsal.

\parahead{Receipts}
Every physical action returns a receipt.
The receipt states whether the action succeeded, the pose that was actually reached, and whether an object is held.
It also reports the deviation from the target and a reason when the action failed.
The model therefore learns from what happened, not from what it commanded.

\subsection{Skill Building}
\label{sec:app_skill_building}

\parahead{Coverage of the demonstrations}
Skill building starts from the demonstrations of all tasks.
Each demonstration is cut at its gripper events into segments of seven kinds, namely grasp, place, carry, touch, release, move, and return home.
A segment is explained when an existing entry point or written skill, with parameters fitted to the demonstration, reproduces it from the state where the segment starts.
Each task is checked in two ways.
Every segment is replayed from its own start, and the plan of the whole task is also rolled out once from the reset, so that each call starts where the previous calls left the scene.
Segments that no entry explains are gaps, and gaps of the same kind and failure class are grouped into clusters.

\parahead{Written skills}
Each cluster is handed to a coding agent that works in a sandbox with fixed budgets of case runs, wall-clock time, and tokens.
A written skill has two parts.
The action part runs at the deployable tier, exactly as the agent runs it in rehearsal and deployment, and it uses only the action API, the object handles it receives, and its parameters.
The fitting part runs only during skill building and may read the recording and the ground-truth scene.
It returns the parameters that reproduce one segment.
These parameters hold no absolute position, since the fit is repeated on the demonstration moved across the table and a parameter that moves with it by more than 1 cm rejects the fit.

\parahead{Gate}
The driver accepts a skill only through its own checks and never relies on the report of the coding agent.
A static check rejects imports outside an allowed set, task names, object labels, and position literals.
The skill must then explain the gaps it was written for.
A transfer check runs its fitted parameters on the same segment of up to three other demonstrations whose object lies at least 5 cm away, and rejects a skill that explains none of them, so that a replay cannot pass as a skill.
Finally, all sweeps of the round run again with the accepted skills and the candidates, and a candidate is dropped when its confirmed losses outweigh its gains.
Every process that imports a written skill runs in the sandbox.

\parahead{Stopping and freezing}
Rounds repeat until coverage stops growing, measured by explained segments, task rollouts that reach the end, official successes of the rolled-out plans, and completed stages.
The accepted skills are then frozen as the skill set of the environment.
The task plans that the builder uses to measure coverage stay in its records and never enter the skill set, a Task Memory, or any input of learning or deployment.

\subsection{Perception}
\label{sec:app_perception}

Perception separates geometry from semantics.
Positions, sizes, and orientations come from the depth images and the calibrated cameras, and foundation models are asked only which object or part is which.
At the start of an episode, the agent identifies the objects of the task once, and an identity is accepted only when a majority of independent detectors agree.
An ambiguous result is reported to the agent instead of resolved silently.
During the episode, objects are tracked through the grasp relations that the actions record instead of being detected again, which keeps their identities stable while the arms occlude them.
Points that belong to the robot are removed with its kinematic model.

\subsection{Task Learning}
\label{sec:app_learning}

\parahead{Imitation}
S1 involves three roles.
A demonstration reader studies the recorded demonstration, including its frames, the motion of the robot, and the key operations extracted from gripper events.
It writes the first version of the Task Memory.
An acting role attempts the task on the instance of the demonstration.
Some attempts may consult the demonstration as hints, and the others rely on the memory alone.
A reviewing role reads the record of each attempt and revises the memory.
An attempt passes when the official success check fires and the key operations of the demonstration were carried out in order.
S1 ends once the attempts without hints pass consistently within a window of recent attempts.
\tabref{tab:app_hparams} lists the values used in all runs.

\parahead{Rehearsal}
S2 runs on new training instances in order, and only the first attempt on each instance counts.
If it fails, the reviewing role edits a cached copy of the memory, and the instance is retried with that copy.
The edits enter the shared memory only if a retry succeeds.
The number of edits allowed per revision and the step size of numerical updates both decrease as rehearsal proceeds, so the memory changes quickly at first and stabilizes later.
Rehearsal stops when enough recent first attempts succeed within a sliding window, or after a fixed instance budget.
If the budget is reached before the early-stopping rule triggers, the version with the best record of first attempts is frozen.

\begin{table}[t]
  \tablestyle{4pt}{1.15}
  \caption{\textbf{Hyperparameters.} Values used in all runs. $k$ counts the rehearsal instances of a task.}
  \label{tab:app_hparams}
  \begin{tabularx}{\linewidth}{ll>{\raggedright\arraybackslash}X}
    \toprule
    Stage & Parameter & Value \\
    \midrule
    S1 & Exit rule & 3 of the 5 most recent attempts without hints pass \\
    S1 & Hints & First attempt, then one after every 2 consecutive failures without hints \\
    S1 & Attempt cap & 60, including attempts with hints \\
    S2 & Counted-instance cap & 60 (20 in the ablations) \\
    S2 & Tries per instance & 3, only the first counts \\
    S2 & Early stop & 3 first-try successes among the 5 most recent counted instances \\
    S2 & Step size at instance $k$ & $\alpha_k = \max(0.2,\ 10/(9+k))$ \\
    S2 & Edit budget at instance $k$ & $\max(1,\ \mathrm{round}(4\alpha_k))$ \\
    All & Reasoning effort & High for every role \\
    \bottomrule
  \end{tabularx}
\end{table}
 
\parahead{Task Memory}
The Task Memory is a structured Markdown document.
Its sections hold the instruction, the roles of the objects and how to find them, the workflow of each stage with its completion criteria, lessons, and notes.
A separate table holds typed parameters such as grasp and placement specifications.
Reviewing roles change the memory only through typed edits, each with a stated reason, and every version is stored and hashed.
Before a version is committed, a linter rejects simulator names, evaluator terms, instance identifiers, absolute positions, distances measured from the robot, and fixed assignments of arms.
The memory thus stays semantic and independent of any instance.

\subsection{Completion Check}
\label{sec:app_completion}

Before the agent may declare a task done, it checks the scene against the final state of the demonstration.
The check uses the completion criteria in the memory and a visual comparison with the final frames of the demonstration.
Only the success check of the simulator credits an episode.

\subsection{Information Boundary}
\label{sec:app_boundary}
\begin{table}[t]
  \tablestyle{4pt}{1.15}
  \caption{\textbf{Data available to each role.} D denotes the deployable tier and P the privileged tier.
  Every role that acts on a new instance uses the deployable tier only.}
  \label{tab:app_roles}
  \begin{tabular}{llcllc}
    \toprule
    Stage & Role & Tier & Demonstration & Privileged data & Edits memory \\
    \midrule
    S1 & Demonstration reader & P & full & tools and demonstration state & yes \\
    S1 & Acting agent & P & as hints only & tools & no \\
    S1 & Reviewing role & P & full & diagnosis and demonstration state & yes \\
    S2 & Acting agent & D & final state & none & no \\
    S2 & Reviewing role & D + diagnosis & none & diagnosis & cached copy \\
    P3 & Acting agent & D & final state & none & no \\
    \bottomrule
  \end{tabular}
\end{table}
 
\parahead{Two tiers}
\tabref{tab:app_roles} lists the data available to each role.
The deployable tier contains RGB-D images from the head and wrist cameras, camera intrinsics and extrinsics, measured and commanded joint states, gripper openings, feedback from the motion planner, the step count, and the instruction.
The privileged tier contains ground-truth object poses, asset annotations and meshes, articulation states, contacts, ground-truth segmentation, the success check, and expert trajectories.

\parahead{Enforcement}
The boundary is enforced by construction.
Agent-side code reaches the simulator only through an interface that returns deployable data, and only the control plane holds the privileged interface.
The hub derives the tools of each role from its tier and refuses privileged calls in evaluation stages.
Deployment mounts no privileged tool at all, and observations, error messages, and tool outputs are filtered so that no privileged value passes through them.
Automated tests check that every evaluation stage refuses privileged tools, that a single module alone may access privileged state, and that no text shown to the model in rehearsal or deployment is tied to a particular task.

\parahead{Supervision during learning}
Privileged data serves only as supervision during learning.
Every role in imitation runs at the privileged tier, since imitation takes place on the instance of the demonstration.
In rehearsal, the reviewing role receives a ground-truth diagnosis of each failure, stated in terms of the objects and roles of the episode.
This supervision reaches the acting agent only through the memory, which must pass the linter.
A privilege-free variant that removes all of this supervision is evaluated in \cref{sec:exp_ablation}.

\section{Additional Quantitative Results}
\label{sec:app_additional_results}

\parahead{Statistical uncertainty}
Over the 1,000 deployment episodes of each model, the 95\% Wilson interval of the mean is 81.0 to 85.6 for GPT-6.1-Sol, 75.2 to 80.4 for GPT-5.6-Luna, and 75.4 to 80.6 for DeepSeek-V4.1-Flash.
Resampling tasks instead of episodes gives 77.5 to 89.0, 70.7 to 84.6, and 71.4 to 84.4, which reflects the spread of difficulty across tasks.
The frozen GPT-6.1-Sol memories, evaluated on four further disjoint seed blocks in the clean setting, reach 83.4\%, 84.6\%, 85.2\%, and 83.0\%.

\parahead{S1 exit}
DeepSeek-V4.1-Flash passes the S1 exit rule on all 50 tasks.
GPT-6.1-Sol passes on 49 (\texttt{place\_fan} does not exit).
GPT-5.6-Luna passes on 47 (\texttt{place\_fan}, \texttt{put\_object\_cabinet}, and \texttt{place\_cans\_plasticbox} do not exit).
A task that does not exit is frozen with the best available memory, defined as the memory version that has at least one official success and the highest Wilson 95\% lower bound on first-try success.
In this run, every non-exiting task has at least one such version, so none scores 0 by default.

\begin{figure}[t]
  \centering
  \includegraphics[width=\textwidth]{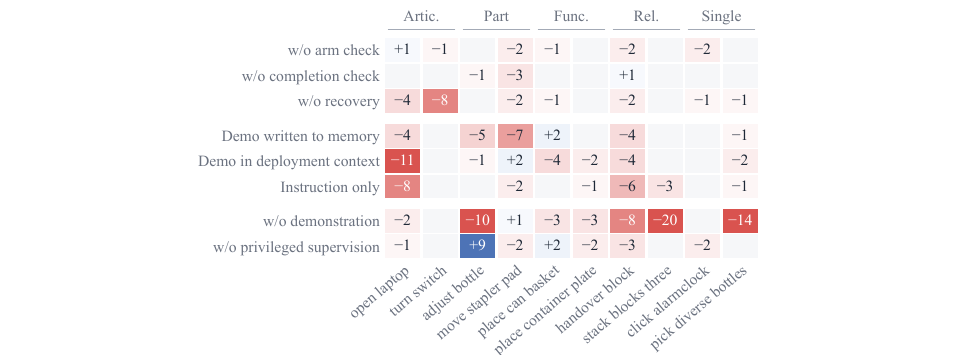}
  \caption{\textbf{Paired changes per task in the ablations.} Each cell is the change in successes over 20 episodes on the same seeds, with blue a gain and red a loss.
  Each variant is compared with the control of its group in \cref{tab:ablation}, where variants without the world are compared with the frozen memory of the main run.
  Tasks that do not converge in rehearsal without the demonstration count as failures.}
  \label{fig:app_paired}
\end{figure}

\parahead{Ablation details}
\tabref{tab:ablation} uses 10 tasks that cover the five categories of \tabref{tab:app_per_task}.
The tasks are \texttt{adjust\_bottle}, \texttt{click\_alarmclock}, \texttt{handover\_block}, \texttt{move\_stapler\_pad}, \texttt{open\_laptop}, \texttt{pick\_diverse\_bottles}, \texttt{place\_can\_basket}, \texttt{place\_container\_plate}, \texttt{stack\_blocks\_three}, and \texttt{turn\_switch}.
All groups use evaluation seeds that are disjoint from all learning and rehearsal seeds.
Without the world, the first variant writes the memory from the same demonstration package as imitation but makes no call to the simulator.
The second deploys a memory that holds only the instruction and shows the deploying agent the demonstration frames and robot keypoints, without its ground-truth state.
The third deploys the instruction-only memory alone.
The privilege-free variant runs every role of imitation at the deployable tier, reads no ground-truth state of the demonstration, and gives the reviewing role no diagnosis.
The learning group caps rehearsal at 20 counted instances per task.
The 11 harder tasks used to evaluate rehearsal are \texttt{hanging\_mug}, \texttt{move\_can\_pot}, \texttt{open\_microwave}, \texttt{pick\_dual\_bottles}, \texttt{place\_bread\_skillet}, \texttt{place\_dual\_shoes}, \texttt{place\_fan}, \texttt{put\_bottles\_dustbin}, \texttt{put\_object\_cabinet}, \texttt{rotate\_qrcode}, and \texttt{stack\_bowls\_three}.
\figref{fig:app_paired} shows the paired changes per task.

\parahead{Unseen object instances}
In 39 tasks the objects vary across instances, and 78\% of the evaluation episodes of these tasks use object instances different from the demonstration.
Within each task, these episodes show no significant difference in success from episodes with the demonstrated instances (Cochran--Mantel--Haenszel $p$ of 0.97, 0.50, and 0.97 for GPT-6.1-Sol, GPT-5.6-Luna, and DeepSeek-V4.1-Flash).
In eight tasks, no evaluation episode uses the demonstrated instances, and success still ranges from 75\% to 100\%.

\parahead{Failure attribution}
\begin{table}[t]
  \tablestyle{4pt}{1.15}
  \caption{\textbf{Failure attribution.} Failed evaluation episodes by physical root cause, attributed from the full event log and action receipts.
  False acceptance counts episodes whose own completion check accepted a state that the official check rejects, and cuts across the three causes.
  Luna's clean execution count includes 2 tool timeouts.}
  \label{tab:app_failure}
  \begin{tabular}{l*{4}{x{46}}}
    \toprule
    & \multicolumn{2}{c}{GPT-5.6-Luna} & \multicolumn{2}{c}{GPT-6.1-Sol} \\
    \cmidrule(lr){2-3} \cmidrule(lr){4-5}
    Cause & Clean & Rand. & Clean & Rand. \\
    \midrule
    Perception & 9 & 13 & 22 & 17 \\
    Knowledge & 28 & 20 & 37 & 38 \\
    Execution & 64 & 87 & 23 & 29 \\
    \midrule
    Total & 101 & 120 & 82 & 84 \\
    False acceptance & 17 (17\%) & 15 (13\%) & 23 (28\%) & 24 (29\%) \\
    \bottomrule
  \end{tabular}
\end{table}
 Each failed evaluation episode is labeled with the earliest unrecovered root cause that made official success unreachable.
We use three categories.
Perception failures misidentify an object or mislocate a part.
Knowledge failures arise when the agent executes the workflow correctly but the stored specification does not match the scene or the instruction.
Execution failures include unreachable targets, empty grasps, dropped objects, and running out of steps during recovery.
The reviewer also records false acceptances, where the agent's completion check accepts a state that the official check rejects.
\tabref{tab:app_failure} reports the full attribution for GPT-6.1-Sol and GPT-5.6-Luna.
To check that the labels do not depend on the labeling model, a different model independently relabels 30 failed episodes.
The two agree on the root cause in 26 of 30 episodes (Cohen's $\kappa=0.82$), and on another 30 failed episodes the distribution of causes from the second model matches that of the first.

\parahead{Cost breakdown}
\begin{table}[t]
  \tablestyle{4pt}{1.15}
  \caption{\textbf{Cost per episode.} Means over the acting agent's episodes in each stage, where S1 and S2 episodes are imitation and rehearsal attempts.
  Input tokens include cached input, with its share in parentheses.
  USD uses each model's standard list API price in October 2026, and the off-peak rate of DeepSeek-V4.1-Flash is half of it.
  Reading the demonstration and reviewing attempts are excluded.
  Attempts redone after infrastructure failures are excluded.}
  \label{tab:app_cost}
  \begin{tabular}{llrrrrr}
    \toprule
    Model & Stage & Episodes & Model calls & Input tokens (M) & Output tokens (k) & USD \\
    \midrule
    GPT-6.1-Sol & S1 & 329 & 22.6 & 0.73 (84\%) & 4.2 & 0.34 \\
     & S2 & 1,397 & 18.4 & 0.52 (86\%) & 3.0 & 0.22 \\
     & P3 & 1,000 & 54.7 & 2.32 (78\%) & 13.4 & 1.33 \\
    \midrule
    GPT-5.6-Luna & S1 & 518 & 27.4 & 0.88 (80\%) & 11.7 & 0.06 \\
     & S2 & 1,732 & 12.0 & 0.33 (82\%) & 3.8 & 0.02 \\
     & P3 & 1,000 & 62.2 & 2.58 (73\%) & 28.3 & 0.21 \\
    \midrule
    DeepSeek-V4.1-Flash & S1 & 301 & 18.2 & 0.94 (93\%) & 61.7 & 0.10 \\
     & S2 & 1,205 & 17.0 & 0.94 (89\%) & 88.7 & 0.14 \\
     & P3 & 1,000 & 53.1 & 3.64 (94\%) & 252.2 & 0.39 \\
    \bottomrule
  \end{tabular}
\end{table}
 \tabref{tab:app_cost} reports, for each foundation model and stage, the number of episodes, the mean number of model calls, and the mean numbers of input and output tokens.
Attempts redone after infrastructure failures are excluded in every stage.
A deployment episode takes a median of 6.4 minutes with GPT-6.1-Sol, 10.4 with GPT-5.6-Luna, and 3.6 with DeepSeek-V4.1-Flash, and about 80\% of this time is spent waiting for the foundation model.
The simulator, the motion planner, and the vision models run on 16 NVIDIA RTX 4090 GPUs across 2 servers, and the foundation models are accessed through their provider APIs.

\begin{table}[p]
  \tablestyle{3.6pt}{0.9}
  \footnotesize
  \caption{\textbf{Per-task success rate (\%) on RoboTwin~2.0.} Tasks are grouped by the five categories of \cref{sec:exp_setup}, with the number of tasks in parentheses.
  \method uses 10 evaluation episodes per task and setting, and $\pi_{0.5}$ is taken from the official leaderboard with 100 episodes.
  C and R denote the clean and randomized settings, and each category ends with its mean.}
  \label{tab:app_per_task}
  \begin{tabular}{l*{8}{x{24}}}
    \toprule
    \multirow{2}{*}{Task} & \multicolumn{2}{c}{$\pi_{0.5}$} & \multicolumn{2}{c}{\method (Flash)} & \multicolumn{2}{c}{\method (Luna)} & \multicolumn{2}{c}{\method (Sol)} \\
    \cmidrule(lr){2-3} \cmidrule(lr){4-5} \cmidrule(lr){6-7} \cmidrule(lr){8-9}
    & C & R & C & R & C & R & C & R \\
    \midrule
    \multicolumn{9}{l}{\textit{Articulated objects (4)}} \\
    \quad\texttt{open\_laptop} & 93 & 68 & 60 & 60 & 90 & 100 & 90 & 100 \\
    \quad\texttt{open\_microwave} & 86 & 59 & 100 & 100 & 100 & 100 & 100 & 100 \\
    \quad\texttt{put\_object\_cabinet} & 39 & 22 & 60 & 40 & 30 & 10 & 90 & 80 \\
    \quad\texttt{turn\_switch} & 52 & 25 & 100 & 100 & 90 & 90 & 100 & 90 \\
    \quad\textit{Mean} & \textit{67.5} & \textit{43.5} & \textit{80.0} & \textit{75.0} & \textit{77.5} & \textit{75.0} & \textit{95.0} & \textit{92.5} \\
    \midrule
    \multicolumn{9}{l}{\textit{Oriented parts (11)}} \\
    \quad\texttt{adjust\_bottle} & 99 & 77 & 80 & 90 & 70 & 70 & 70 & 80 \\
    \quad\texttt{beat\_block\_hammer} & 93 & 23 & 90 & 80 & 90 & 80 & 90 & 90 \\
    \quad\texttt{hanging\_mug} & 17 & 14 & 0 & 0 & 30 & 0 & 30 & 20 \\
    \quad\texttt{move\_stapler\_pad} & 20 & 18 & 100 & 90 & 70 & 90 & 90 & 100 \\
    \quad\texttt{place\_dual\_shoes} & 70 & 45 & 60 & 80 & 70 & 60 & 80 & 80 \\
    \quad\texttt{place\_fan} & 66 & 26 & 10 & 20 & 30 & 30 & 40 & 70 \\
    \quad\texttt{place\_mouse\_pad} & 34 & 31 & 100 & 100 & 100 & 100 & 100 & 90 \\
    \quad\texttt{place\_phone\_stand} & 65 & 37 & 90 & 100 & 70 & 80 & 80 & 90 \\
    \quad\texttt{place\_shoe} & 82 & 50 & 90 & 80 & 90 & 90 & 80 & 80 \\
    \quad\texttt{rotate\_qrcode} & 90 & 21 & 70 & 30 & 70 & 60 & 30 & 60 \\
    \quad\texttt{scan\_object} & 49 & 40 & 0 & 30 & 30 & 20 & 60 & 60 \\
    \quad\textit{Mean} & \textit{62.3} & \textit{34.7} & \textit{62.7} & \textit{63.6} & \textit{65.5} & \textit{61.8} & \textit{68.2} & \textit{74.5} \\
    \midrule
    \multicolumn{9}{l}{\textit{Functional points on a receiver (10)}} \\
    \quad\texttt{place\_bread\_basket} & 63 & 60 & 90 & 80 & 80 & 70 & 100 & 100 \\
    \quad\texttt{place\_bread\_skillet} & 76 & 47 & 30 & 50 & 20 & 20 & 60 & 60 \\
    \quad\texttt{place\_burger\_fries} & 82 & 86 & 90 & 80 & 100 & 100 & 100 & 90 \\
    \quad\texttt{place\_can\_basket} & 56 & 10 & 60 & 80 & 70 & 70 & 70 & 40 \\
    \quad\texttt{place\_cans\_plasticbox} & 31 & 73 & 90 & 80 & 100 & 80 & 100 & 100 \\
    \quad\texttt{place\_container\_plate} & 92 & 72 & 90 & 90 & 100 & 60 & 90 & 90 \\
    \quad\texttt{place\_empty\_cup} & 96 & 80 & 90 & 80 & 100 & 90 & 100 & 90 \\
    \quad\texttt{place\_object\_basket} & 70 & 29 & 80 & 90 & 60 & 80 & 90 & 80 \\
    \quad\texttt{place\_object\_scale} & 78 & 38 & 80 & 90 & 100 & 90 & 100 & 100 \\
    \quad\texttt{place\_object\_stand} & 85 & 47 & 90 & 90 & 80 & 100 & 90 & 80 \\
    \quad\textit{Mean} & \textit{72.9} & \textit{54.2} & \textit{79.0} & \textit{81.0} & \textit{81.0} & \textit{76.0} & \textit{90.0} & \textit{83.0} \\
    \midrule
    \multicolumn{9}{l}{\textit{Relative placement (10)}} \\
    \quad\texttt{handover\_block} & 30 & 12 & 70 & 60 & 90 & 80 & 70 & 50 \\
    \quad\texttt{move\_can\_pot} & 60 & 10 & 80 & 60 & 50 & 40 & 50 & 20 \\
    \quad\texttt{move\_pillbottle\_pad} & 46 & 32 & 90 & 90 & 100 & 80 & 90 & 100 \\
    \quad\texttt{place\_a2b\_left} & 71 & 49 & 100 & 90 & 100 & 90 & 100 & 100 \\
    \quad\texttt{place\_a2b\_right} & 68 & 39 & 100 & 100 & 100 & 90 & 100 & 100 \\
    \quad\texttt{stack\_blocks\_three} & 77 & 25 & 60 & 80 & 100 & 100 & 100 & 100 \\
    \quad\texttt{stack\_blocks\_two} & 91 & 43 & 100 & 100 & 100 & 100 & 100 & 100 \\
    \quad\texttt{stack\_bowls\_three} & 81 & 52 & 40 & 70 & 20 & 20 & 30 & 50 \\
    \quad\texttt{stack\_bowls\_two} & 94 & 70 & 90 & 70 & 90 & 80 & 90 & 90 \\
    \quad\texttt{stamp\_seal} & 53 & 21 & 70 & 90 & 90 & 80 & 100 & 100 \\
    \quad\textit{Mean} & \textit{67.1} & \textit{35.3} & \textit{80.0} & \textit{81.0} & \textit{84.0} & \textit{76.0} & \textit{83.0} & \textit{81.0} \\
    \midrule
    \multicolumn{9}{l}{\textit{Single objects (15)}} \\
    \quad\texttt{blocks\_ranking\_rgb} & 72 & 44 & 80 & 70 & 90 & 80 & 100 & 100 \\
    \quad\texttt{blocks\_ranking\_size} & 45 & 21 & 70 & 100 & 100 & 100 & 100 & 100 \\
    \quad\texttt{click\_alarmclock} & 65 & 50 & 100 & 90 & 90 & 90 & 100 & 100 \\
    \quad\texttt{click\_bell} & 31 & 35 & 100 & 100 & 100 & 100 & 100 & 100 \\
    \quad\texttt{dump\_bin\_bigbin} & 92 & 84 & 70 & 60 & 70 & 60 & 70 & 70 \\
    \quad\texttt{grab\_roller} & 99 & 97 & 100 & 100 & 90 & 100 & 100 & 100 \\
    \quad\texttt{handover\_mic} & 98 & 6 & 90 & 100 & 100 & 100 & 100 & 100 \\
    \quad\texttt{lift\_pot} & 98 & 35 & 100 & 90 & 100 & 100 & 100 & 90 \\
    \quad\texttt{move\_playingcard\_away} & 84 & 65 & 90 & 100 & 100 & 100 & 100 & 100 \\
    \quad\texttt{pick\_diverse\_bottles} & 66 & 42 & 30 & 70 & 30 & 50 & 50 & 30 \\
    \quad\texttt{pick\_dual\_bottles} & 72 & 57 & 90 & 100 & 100 & 100 & 40 & 60 \\
    \quad\texttt{press\_stapler} & 79 & 63 & 100 & 100 & 90 & 90 & 100 & 100 \\
    \quad\texttt{put\_bottles\_dustbin} & 78 & 50 & 70 & 30 & 50 & 40 & 60 & 80 \\
    \quad\texttt{shake\_bottle} & 100 & 100 & 100 & 90 & 100 & 100 & 100 & 100 \\
    \quad\texttt{shake\_bottle\_horizontally} & 99 & 100 & 100 & 100 & 100 & 90 & 100 & 100 \\
    \quad\textit{Mean} & \textit{78.5} & \textit{56.6} & \textit{86.0} & \textit{86.7} & \textit{87.3} & \textit{86.7} & \textit{88.0} & \textit{88.7} \\
    \midrule
    Average & 70.7 & 46.0 & 77.8 & 78.4 & 79.8 & 76.0 & 83.6 & 83.2 \\
    \bottomrule
  \end{tabular}
\end{table}
 
\section{Additional Qualitative Results}
\label{sec:app_qualitative}

\parahead{Task Memory structure}
\figref{fig:app_memory} shows the frozen Task Memory of \texttt{place\_object\_basket} as the actor receives it at evaluation.
The memory is organized into five sections.
The instruction and object roles describe the task and how to identify each object.
The workflow lists eight stages in order, each with its goal, the arm to use, the API calls, and the condition that marks it as done.
The parameter table holds 21 typed specifications, including grasp poses, placement offsets, and scalar thresholds.
Lessons record failure patterns that the reviewing role identified during rehearsal.
In this example, rehearsal added stage 3 (a handover to the other arm) and both lessons, shown in blue.
The text shown is quoted from the stored file and represents the full context the acting agent receives at every step during evaluation.

\parahead{Context that changes the outcome}
\figref{fig:app_context} shows two cases where a single piece of learned context changes the evaluation outcome on the same seeds.

In (a), rehearsal adds one lesson to the Task Memory of \texttt{move\_stapler\_pad}.
Without this lesson, the agent finishes on xy overlap alone and the official check rejects the episode.
With the lesson, the agent verifies containment before finishing.
On the same clean seeds, success rises from 4 of 10 to 9 of 10.

In (b), the reflection loop writes two guidance entries for \texttt{handover\_block} over three rounds.
The first entry (added in round 1) corrects a misinterpretation of abbreviated wording that causes the agent to stop after the donor hold.
The second entry (added in round 3) tells the agent to verify placement against the visible pad edges and recenter if displaced.
With both entries, success on the same seeds rises from 4 of 10 to 10 of 10.
These examples illustrate that the WICL loop writes targeted, readable corrections into the context and that small semantic changes can recover multiple episodes.

\definecolor{ctxadd}{HTML}{2E5AA8}
\definecolor{ctxdel}{HTML}{E0793A}
\definecolor{ctxmeta}{HTML}{627D95}
\newtcolorbox{ctxbox}[1]{enhanced, colback=abstractbg, colframe=abstractbg, coltitle=bannerblue, boxrule=0pt,
  arc=2pt, left=5pt, right=5pt, top=3pt, bottom=4pt, fonttitle=\bfseries\small, title={#1},
  attach title to upper, after title={\par\smallskip}, fontupper=\scriptsize\ttfamily}
\newcommand{\memsec}[2]{\textsf{\textbf{\color{bannerblue}#1}}\par\vspace{1pt}{\scriptsize\color{ctxmeta}#2}}
\newcommand{\memcode}[1]{{\fontsize{6.6}{8}\selectfont\ttfamily #1}}
\newcommand{\memadd}[1]{\textcolor{ctxadd}{#1}}
\begin{figure}[!t]
  \centering
  \begin{tcolorbox}[enhanced, colback=abstractbg, colframe=abstractbg, boxrule=0pt, arc=2pt,
      left=6pt, right=6pt, top=4pt, bottom=4pt]
    \footnotesize\setlength{\tabcolsep}{0pt}\renewcommand{\arraystretch}{1.0}
    \arrayrulecolor{white}
    \begin{tabularx}{\linewidth}{@{}>{\raggedright\arraybackslash}p{0.17\linewidth}@{\hspace{8pt}}X@{}}
      \memsec{INSTRUCTION}{from the demonstration} &
      Grab the bright sleek toycar, drop it in the basket with perforated sides, then move the basket with perforated sides. \\[3pt] \specialrule{1pt}{0pt}{3pt}
      \memsec{OBJECTS}{role, description, $x$ axis} &
      \memcode{toycar}: the object to drop into the container; a small rigid toy vehicle, grasped across its body [...]. $x$ toward the front nose. \newline
      \memcode{basket}: receives the object and is then moved while containing it [...]. $x$ toward the raised handle crossbar end. \\[3pt] \specialrule{1pt}{0pt}{3pt}
      \memsec{WORKFLOW}{stages with goal, arm, calls, done} &
      \memcode{1 prepare} $\rightarrow$ \memcode{2 grasp\_toycar} $\rightarrow$ \memadd{\memcode{3 handover\_toycar}} $\rightarrow$ \memcode{4 lift\_toycar} $\rightarrow$ \memcode{5 drop\_toycar} $\rightarrow$ \memcode{6 retreat} $\rightarrow$ \memcode{7 grasp\_basket} $\rightarrow$ \memcode{8 move\_basket} \par\vspace{2pt}
      \memcode{5 drop\_toycar:} Move above the basket opening with the recorded tilt and release the toycar so it falls into the basket. [...] \newline
      \memcode{call: place\_actor(arm=<arm>, obj=toycar, place=\{toycar\_drop\}, pre\_dis=\{drop\_pre\_dis\}, dis=\{drop\_dis\}, is\_open=true, constrain='align'); [...]} \newline
      \memcode{done: gripper\_open(<arm>, \{open\_fraction\}) and not holding(<arm>)} \\[3pt] \specialrule{1pt}{0pt}{3pt}
      \memsec{PARAMETERS}{fitted from the demonstration} &
      \memcode{toycar\_drop = place\_spec(target=basket top\_center, frame=target, offset\_m=[0.0781, 0.0067, 0.0234], orient=relative, rpy\_deg=[-42.7, 84.1625, -138.2], functional=bottom, [...])} \newline
      \memcode{lift\_height = 0.1 m} \quad [...] 19 more grasp, path, place, and done specifications \\[3pt] \specialrule{1pt}{0pt}{3pt}
      \memsec{LESSONS}{written from failures} &
      \memadd{\memcode{L2} \textit{When} the object has been released with the fitted tilted placement and the empty gripper must clear the basket,
      \textit{do} keep the arm selected for the object through retreat and use \memcode{back\_to\_origin(arm=<arm>)} for the empty gripper, allowing a planned path and orientation change.
      Do not replace this with a long straight world displacement at the release orientation, even if its endpoint appears clear. [...]} \newline
      \memadd{\memcode{L1}} [...] \\[3pt] \specialrule{1pt}{0pt}{3pt}
      \memsec{NOTES}{free-text cautions} &
      The demonstration ends with the loaded basket still grasped, not set down. [...]
    \end{tabularx}
  \end{tcolorbox}
  \caption{\textbf{Task Memory.} The frozen Task Memory of \texttt{place\_object\_basket} learned by GPT-6.1-Sol, as the actor receives it at evaluation.
  The left column names each section and what it holds. Rehearsal added stage 3, a handover to the other arm, and both lessons, shown in blue.
  Text is quoted from the stored file, and omitted text is marked [...].}
  \label{fig:app_memory}
\end{figure}
 \begin{figure}[t]
  \centering
  \begin{tcolorbox}[enhanced, colback=abstractbg, colframe=abstractbg, boxrule=0pt, arc=2pt,
      left=6pt, right=6pt, top=4pt, bottom=4pt]
    \footnotesize\setlength{\tabcolsep}{0pt}\renewcommand{\arraystretch}{1.0}
    \arrayrulecolor{white}
    \begin{tabularx}{\linewidth}{@{}>{\raggedright\arraybackslash}p{0.2\linewidth}@{\hspace{8pt}}X@{}}
      \memsec{(a) LESSON}{added in rehearsal to the Task Memory of \texttt{move\_stapler\_pad}}
      \par\vspace{3pt}{\scriptsize Clean, same seeds}\par\textbf{4/10 $\rightarrow$ 9/10} &
      \memcode{L1} \textit{When} grounding the stapler or verifying its released placement, especially when \memcode{check\_done} passes but \memcode{on} or official completion fails,
      \textit{do} use the live instruction's stapler and mat descriptions and \memcode{choose\_arm} rule.
      Confirm the whole stapler includes its lower base, upper body and hinge before applying the pose grasp.
      After release, withdraw and remeasure the complete object from unobstructed views before checking \memcode{on} and \memcode{check\_done}.
      Never finish on xy overlap alone: if \memcode{on} fails or finish is refused, verify the measured base/contact and containment before a reseating correction; do not repeatedly nudge from a stale or partial box. \\[3pt]
      \specialrule{1pt}{0pt}{3pt}
      \memsec{(b) GUIDANCE}{written by reflection for \texttt{handover\_block}}
      \par\vspace{3pt}{\scriptsize Round 3, same seeds}\par\textbf{4/10 $\rightarrow$ 10/10} &
      \memcode{T-handover\_block-02} \textcolor{ctxmeta}{(active; helpful 8, harmful 2; added r1)} \newline
      \textit{Watch for} abbreviated grab wording is interpreted as requiring a permanent donor hold, despite the empty blue pad and unfinished handover
      $\rightarrow$ Treat abbreviated grab wording as naming the donor stage, not the final state. If the pad is empty after grasping, continue the handover and released pad placement rather than stopping with the donor hold. \par\vspace{3pt}
      \memcode{T-handover\_block-03} \textcolor{ctxmeta}{(active; helpful 6, harmful 0; added r3)} \newline
      \textit{Instead of} Task Memory step \memcode{place\_on\_pad}: \textit{When} the receiver is lowering the block onto the pad, whose refined target may disagree with the visible blue edges
      $\rightarrow$ Lower using the stored pad placement without opening; verify the bottom lies inside the visible pad. If displaced, lift and recenter while retaining the receiver hold, then release and perform the stored clearance. [...]
    \end{tabularx}
  \end{tcolorbox}
  \caption{\textbf{Learned context that changes the outcome.} Left, the source of each text and success on the same seeds without and with it.
  (a) The only lesson rehearsal adds to this Task Memory. Without it, all five lost episodes end with a finish on overlap alone.
  (b) Two guidance entries after round 3. Text is quoted, with cuts marked.}
  \label{fig:app_context}
\end{figure}
 
\section{Limitations and Future Work}
\label{sec:app_limitations}

\parahead{Limitations}
\method relies on commercial foundation models served through APIs, whose service quality varies over time and can shift success by several points between sessions.
Inference is also slow compared with a trained policy, since an evaluation episode with GPT-6.1-Sol takes on average 54.7 model calls and the robot waits for each call before it acts.
Reflection improves the frozen memories in every round, but the net gain shrinks quickly as the easier failure modes are addressed first, and sparse success signals make the remaining failures hard to attribute.
All evaluated tasks involve rigid or articulated objects.
Deformable objects such as cloth and cables, and tasks that require tightly coordinated contact between the two arms such as handovers, remain challenging and call for richer perception, skills, and worlds.

\parahead{Future work}
Beyond these limitations, the agentic side of WICL remains largely unexplored.
Agents could build new skills when existing ones fail, coordinate several specialized agents over a long task, or rehearse in learned world models when no simulator is available.
Our results also suggest that agent-centric embodied models are a promising direction.
A foundation model trained to use embodied tools, read physical feedback, and maintain a task memory could make WICL faster, cheaper, and more reliable than general-purpose models accessed through an API.
  
\clearpage

\bibliographystyle{plainnat}
\bibliography{main}

\end{document}